\pdfoutput=1
\documentclass[12pt,amsmath,amssymb,longbibliography,nofootinbib,showkeys,eprint]{revtex4-2}
\usepackage[activate={true,nocompatibility},final,tracking=true,kerning=true,spacing=true]{microtype}
\usepackage{filecontents} 
\usepackage{epigraph}
\usepackage{graphicx}
\usepackage{epstopdf}
\usepackage{braket}
\usepackage{bm}
\usepackage{numprint}
\usepackage{float}
\usepackage[T1,T2A]{fontenc}
\usepackage[utf8]{inputenc}
\usepackage{hhline}
\usepackage{xfrac}
\usepackage{amsthm}
\theoremstyle{definitions}

\usepackage{makecell}
\usepackage{tikz-timing}
\usepackage{dsfont}
\usepackage{calc}
\usepackage{seqsplit}
\usepackage{hyperref}

\allowdisplaybreaks

\begin{document}
\preprint{V.M.}
\title{On Machine Learning Knowledge Representation In The Form
  Of Partially Unitary Operator. Knowledge Generalizing Operator}
\author{Vladislav Gennadievich \surname{Malyshkin}} 
\email{malyshki@ton.ioffe.ru}
\affiliation{Ioffe Institute, Politekhnicheskaya 26, St Petersburg, 194021, Russia}

\date{December, 22, 2022}

\begin{abstract}
\begin{verbatim}
$Id: KnowledgeRepresentationAsUnitaryOperator.tex,v 1.449 2022/12/22 06:08:07 mal Exp $
\end{verbatim}
A new form of ML knowledge representation with
high generalization power is developed
and implemented
\hyperref[numAlgorithm]{numerically}.
Initial \emph{IN} attributes and \emph{OUT} class label
are transformed into the corresponding Hilbert spaces
by considering localized wavefunctions.
A partially unitary operator optimally
converting a state from \emph{IN} Hilbert space into \emph{OUT} Hilbert space
is then built from an optimization problem of transferring
maximal possible probability from  \emph{IN} to \emph{OUT},
this leads to the formulation of a new algebraic problem.
Constructed Knowledge Generalizing Operator
$\mathcal{U}$ can be considered as a \emph{IN} to \emph{OUT}
\hyperref[KGOWithoutContributingSubspace]{quantum channel};
it is a partially unitary rectangular matrix of the dimension
$\dim(\emph{OUT}) \times \dim(\emph{IN})$ transforming
operators  as
$A^{\emph{OUT}}=\mathcal{U} A^{\emph{IN}} \mathcal{U}^{\dagger}$.
Whereas only operator $\mathcal{U}$ projections squared are observable
$\Braket{\emph{OUT}|\mathcal{U}|\emph{IN}}^2$ (probabilities),
the
\hyperref[eigenvaluesLikeProblem]{fundamental equation}
is formulated for the operator $\mathcal{U}$ itself.
This is the reason of high generalizing power of the approach;
the situation is the same as for the Schr\"{o}dinger equation:
we can only measure $\psi^2$,
but the equation is written for $\psi$ itself.
\end{abstract}

\maketitle
\newpage

\section{\label{intro}Introduction}
There are four key elements in any ML approach\cite{maloldarxiv}:
\begin{itemize}
\item Attributes selection.
\item Knowledge representation.
\item Quality criteria (norm).
\item Search algorithm to find the solution in knowledge representation space.
\end{itemize}
Knowledge representation is the most important element as it determines
generalization power of a ML system.
The progress in knowledge representation
from linear regression coefficients,
perceptron weights\cite{rosenblatt1958perceptron},
statistical learning\cite{vapnik1974method,vapnik1974methodP2},
and logical approaches\cite{hajek1977generation}
to support vector machines\cite{vapnik2013nature},
rules and decision trees\cite{witten2002data},
fuzzy logic\cite{zadeh1965fuzzy,hajek1995fuzzy},
and deep learning\cite{bengio2013representation} has been the direction of
ML development within the last two decades. 

These approaches, however, share one common feature
that limits their applicability. All of them
typically construct a norm,
loss function,
penalty function,
metric,
distance function,
etc. on class label (attributes to predict)
difference from the target
and perform it's optimization
on training data.
Selection of the norm is a  complex task, moreover, 
the concept of ``norm'' is of statistical type
and cannot be applied in every situation.
In our earlier works\cite{malyshkin2015norm,bobyl2020radon}
we introduced a ``norm-free'' approach where the norm
was replaced by projection operators. The idea takes
inspiration in quantum mechanics where
the outcomes of an observable $f$ (obtained as an operator's spectrum $\Ket{f|\psi^{[i]}}=\lambda^{[i]}\Ket{\psi^{[i]}}$)
and the probabilities of outcomes are separated;
for a given state $\Ket{\psi}$ the probabilities of $\lambda^{[i]}$
outcomes
are obtained as projections to  $\Ket{\psi^{[i]}}$ eigenvectors
$\Braket{\psi|\psi^{[i]}}^2$. This approach comes in two ``flavors''\cite{malyshkin2019radonnikodym}: interpolatory type (where the outcome is obtained as
regular Radon--Nikodym derivative) and classification type (where the outcome
is obtained as prior weight adjusted Radon--Nikodym derivative, a ``Bayesian'' style).

While these results are of great interest as they overcome
one of the most difficult problem in ML (norm selection) and produce
gauge-invariant solutions,
they, as the other approaches to ML, still have a limitation
in generalization power. The problem with this our approach\cite{malyshkin2019radonnikodym}
is that it is still of ``joint distribution generalization'' type.
Effectively it constructs a joint distribution of
(attributes, class label) pairs and then is trying to generalize from it.
Some ML approaches, such as
statistical learning,
support vector machines,
rules and decision trees,
Bayesian learning, etc.
do this ``joint distribution generalization'' explicitly;
the others, such as neural networks,
hidden Markov model, almost all logic models, etc.
in fact also do a ``joint distribution generalization'',
but do it implicitly.

The problem with  ``joint distribution generalization'' approaches
is that they can only predict the outcomes
that already have corresponding (attributes, class label) observations
in training data. For example if we apply such an approach to
periodic planetary motion -- we obtain an accurate prediction,
but applying it to a
\href{https://en.wikipedia.org/wiki/List_of_hyperbolic_comets}{hyperbolic comet}
would be a failure as the comet only travel through the Solar system once.
However, both (planet and hyperbolic comet) are governed by the same Newtonian laws
and their motion is the phenomena of the same kind.
Newtonian mechanics has a more powerful generalization
than the  ``joint distribution generalization''.

This work is the first work where we go beyond the
``joint distribution generalization'' in ML knowledge representation.

\section{\label{dataAndExamples}Input Data and Simple Models}
Whereas the developed approach can be applied to input data
of various forms, for the purpose of comparison with well known
models we will be considering only the data of supervised learning form\footnote{
\label{aboutProductAttributes}
The data can be possibly ``producted'' to some order $\mathcal{D}$.
For example take $n$ initial  $x_k$
and construct
$x_{\mathbf{k}}= x_0^{k_0}x_1^{k_1}\dots x_{n-1}^{k_{n-1}}$
with multi-index $\mathbf{k}=(k_0,k_1,\dots,k_{n-1})$
subject to $\mathcal{D}=\sum\limits_{j=0}^{n-1}k_j$.
From initial $n$ attributes $x_k$ we now obtained 
$\mathcal{N}(n,\mathcal{D}) = C_{n+\mathcal{D}-1}^{\mathcal{D}}$
attributes $x_{\mathbf{k}}$ producted to the order $\mathcal{D}$,
see \cite{malyshkin2019radonnikodym}.
}:
\begin{align}
  (x_0,x_1,\dots,x_k,\dots,x_{n-1})^{(l)}&\to
  (f_0,f_1,\dots,f_j,\dots,f_{m-1})^{(l)}
  & \text{weight $\omega^{(l)}$}  \label{mlproblemVector} \\
  \mathbf{x}^{(l)} &\to \mathbf{f}^{(l)} 
  \nonumber
\end{align}
where an attributes vector $\mathbf{x}$ of the dimension $n$
is mapped to a class label vector $\mathbf{f}$ of the dimension $m$
for all $l=1\dots M$ observations.
An average
 $\Braket{\cdot}$
is defined as the sum
over all $M$ observations sample:
\begin{align}
\Braket{1}&=\sum\limits_{l=1}^{M}   \omega^{(l)}\label{mudef}\\ 
  \Braket{h(\mathbf{f})g(\mathbf{x})}&=\sum\limits_{l=1}^{M} h(\mathbf{f}^{(l)})g(\mathbf{x}^{(l)}) \omega^{(l)} 
  \label{gfaverage}
\end{align}
Here $h(\mathbf{f})$ and $g(\mathbf{x})$ are some functions
on $\mathbf{f}$ and  $\mathbf{x}$, for example a polynomial
or Christoffel function $K(x)$ from (\ref{ChristoffelFunctionDef}).
In this paper we will be considering the models built on
``moments'' --- some average of a polynomial function on $x_k$ and $f_j$;
an example of such an average is
$\Braket{x_kx_{k^{\prime}}f_jf_{j^{\prime}}}$. As a constant has always to be present
in $\mathbf{x}$ and $\mathbf{f}$ bases the tensor
$\Braket{x_kx_{k^{\prime}}f_jf_{j^{\prime}}}$
includes all lower order averages such as $\Braket{x_k x_{k^{\prime}}}$
and $\Braket{f_j f_{j^{\prime}}}$.
Introduce Gram matrices $G_{kk^{\prime}}^{\mathbf{x}}$ and $G_{jj^{\prime}}^{\mathbf{f}}$
for $\mathbf{x}$-- and $\mathbf{f}$-- spaces respectively:
\begin{align}
G_{kk^{\prime}}^{\mathbf{x}} &= \Braket{x_k x_{k^{\prime}}} \label{GramX} \\
G_{jj^{\prime}}^{\mathbf{f}} &= \Braket{f_j f_{j^{\prime}}} \label{GramF}
\end{align}
We will assume that Gram matrices are non--degenerated,
otherwise a regularization to be applied to
$\mathbf{x}$ and $\mathbf{f}$ bases,
see ``\emph{Appendix A: Regularization Example}'' of \cite{malyshkin2019radonnikodym}.

A few familiar examples. Least squares solution of $\mathbf{f}$ on $\mathbf{x}$
requires Gram matrix $G_{kk^{\prime}}^{\mathbf{x}}$ and $\Braket{f_j x_k}$ moments
as input to obtain $f_j(\mathbf{x})=\sum_{k=0}^{n-1}\beta_k x_k$
as linear system solution:
\begin{align}
&\Braket{\left[f_j-\sum_{k=0}^{n-1}\beta_k x_k\right]^2}\to\min\label{lsqmin}\\
&f_j(\mathbf{x})\approx\sum\limits_{k,k^{\prime}=0}^{n-1} x_kG_{kk^{\prime}}^{\mathbf{x};\,-1} \Braket{f_j x_{k^{\prime}}}
\label{fxapproxLS}
\end{align}
Here $G_{kk^{\prime}}^{\mathbf{x};\,-1}$ is Gram matrix (\ref{GramX}) inverse.
The (\ref{fxapproxLS}) is $m$ different predictors 
each one is applied to it's own class label component $f_j$, $j=0\dots m-1$.
Least squares knowledge representation model has limited predictive power and
low outlier stability
but it is very easy to implement numerically
and obtained solution is gauge-invariant relatively
an arbitrary non--degenerated linear transform of
$\mathbf{x}$ and $\mathbf{f}$:
\begin{subequations}
\label{gaugeXF}
\begin{align}
x^{\prime}_{k}&=\sum\limits_{k^{\prime}=0}^{n-1}T_{kk^{\prime}} x_{k^{\prime}} \label{gaugeX}\\
f^{\prime}_{j}&=\sum\limits_{j^{\prime}=0}^{m-1}T_{jj^{\prime}} f_{j^{\prime}} \label{gaugeF}
\end{align}
\end{subequations}
This often makes the least squares model the first choice
to start data analysis despite all the drawbacks.
The model has the properties similar to
``joint distribution generalization'' on the support of $\Braket{\cdot}$
and typically diverges for $\mathbf{x}$ outside of the support interval;
it has low generalization power.

Radon--Nikodym model consists in constructing
a weight density $\psi^2_{\mathbf{y}}(\mathbf{x})$ localized
at $\mathbf{x}=\mathbf{y}$
and then averaging $\mathbf{f}$ with it:
\begin{align}
       \psi_{\mathbf{y}}(\mathbf{x})&
           =\sqrt{K(\mathbf{y})}\sum\limits_{i,k=0}^{n-1}y_iG^{\mathbf{x};\,-1}_{ik}x_k
           =
           \frac{\sum\limits_{i,k=0}^{n-1}y_iG^{\mathbf{x};\,-1}_{ik}x_k}
           {\sqrt{\sum\limits_{i,k=0}^{n-1}y_iG^{\mathbf{x};\,-1}_{ik}y_k}}             
       =\frac{\sum\limits_{i=0}^{n-1}\psi^{[i]}(\mathbf{y})\psi^{[i]}(\mathbf{x})}
           {\sqrt{\sum\limits_{i=0}^{n-1}\left[\psi^{[i]}(\mathbf{y})\right]^2}}
  \label{psiYlocalized}\\
  K(\mathbf{x})&=\frac{1}{\sum\limits_{i,k=0}^{n-1}x_iG^{\mathbf{x};\,-1}_{ik}x_k}
  =\frac{1}{\sum\limits_{i=0}^{n-1}\left[\psi^{[i]}(\mathbf{x})\right]^2}
  \label{ChristoffelFunctionDef}\\
  f_j(\mathbf{x})&\approx
  \frac{\Braket{\psi^2_{\mathbf{x}}f_j}}
{\Braket{\psi^2_{\mathbf{x}}}}
=
    \frac{\sum\limits_{i,q,s,k=0}^{n-1}x_iG^{\mathbf{x};\,-1}_{iq}\Braket{x_q x_s f_j}G^{\mathbf{x};\,-1}_{sk}x_k}
  {\sum\limits_{i,k=0}^{n-1}x_iG^{\mathbf{x};\,-1}_{ik}x_k}
  =\frac{\sum\limits_{i,k=0}^{n-1}\psi^{[i]}(\mathbf{x})\Braket{\psi^{[i]}|f_j|\psi^{[k]}}\psi^{[k]}(\mathbf{x})}
           {\sum\limits_{i=0}^{n-1}\left[\psi^{[i]}(\mathbf{x})\right]^2}
  \label{RNfsolutionX}
\end{align}
In Eq. (\ref{RNfsolutionX}) the Radon--Nikodym approximation is
presented in two bases: original $x_k$, for which $\Braket{x_ix_k}=G^{\mathbf{x}}_{ik}$,
and in some orthogonalized  basis $\Ket{\psi^{[i]}}$ such that
$\Braket{\psi^{[i]}|\psi^{[k]}}=\delta_{ik}$.
Whereas in least squares approximation (\ref{fxapproxLS})
the $f_j(\mathbf{x})$ is a linear combination of basis function $x_k$,
in the Radon--Nikodym approximation (\ref{RNfsolutionX})
it is a ratio of two quadratic forms on basis function $x_k$
with the matrices
$\sum_{q,s=0}^{n-1}G^{\mathbf{x};\,-1}_{iq}\Braket{x_q x_s f_j}G^{\mathbf{x};\,-1}_{sk}$
and $G^{\mathbf{x};\,-1}_{ik}$.
By construction it is an averaging with positive weight\footnote{
For a given $\psi$ the normalizing condition
is $1=\Braket{\psi^2}$, this is required to properly average
an observable $\Braket{f\psi^2}$.
In applications, however, the number of ``covered''
observations is often also required,
for example to estimate possible data overfitting;
the total coverage is $\Braket{1}$ (\ref{mudef}).
To estimate the number of observations covered by a given
 $\psi$ one can use the Christoffel function $K(\mathbf{x})$ (\ref{ChristoffelFunctionDef})
 to estimate the coverage as: $\mathrm{Coverage}_{\psi}\approx\Braket{K\psi^2}$.
 With an expansion of $K(\mathbf{x})$ in spectrum\cite{malyshkin2019radonnikodym}
 $\Ket{K|\psi^{[i]}}=\lambda^{[i]}\Ket{\psi^{[i]}}$
 one can obtain an expansion ``by coverage'';
this removes the major limitation of the principal components method:
it's dependence
on the scale of $\mathbf{x}$ attributes.
}
$\Braket{\psi^2f_j}\big/\Braket{\psi^2}$ thus the bounds of $f_j$ are preserved
and the approximation (\ref{RNfsolutionX}) tends to a constant
when some $x_k\to\infty$
The calculation requires Gram matrix $G_{kk^{\prime}}^{\mathbf{x}}$
and $\Braket{x_kx_{k^{\prime}}f_j}$ moments
as input (compare with
$G_{kk^{\prime}}^{\mathbf{x}}$
and $\Braket{x_kf_j}$ required for least squares
$f(\mathbf{x})\approx\Braket{\psi_{\mathbf{x}}|f}\psi_{\mathbf{x}}(\mathbf{x})=\sum_{i,k=0}^{n-1}x_iG^{\mathbf{x};\,-1}_{ik}\Braket{x_kf}$ approximation);
the result is gauge--invariant relatively (\ref{gaugeXF}).
The (\ref{RNfsolutionX}) is the solution of ``interpolatory''
type as it does not take into account ``prior probabilities'',
see \cite{malyshkin2019radonnikodym} for ``classification'' type  solution
with prior probabilities taken into account,
a ``Bayesian style''.

\begin{figure}[t]
  \includegraphics[width=16cm]{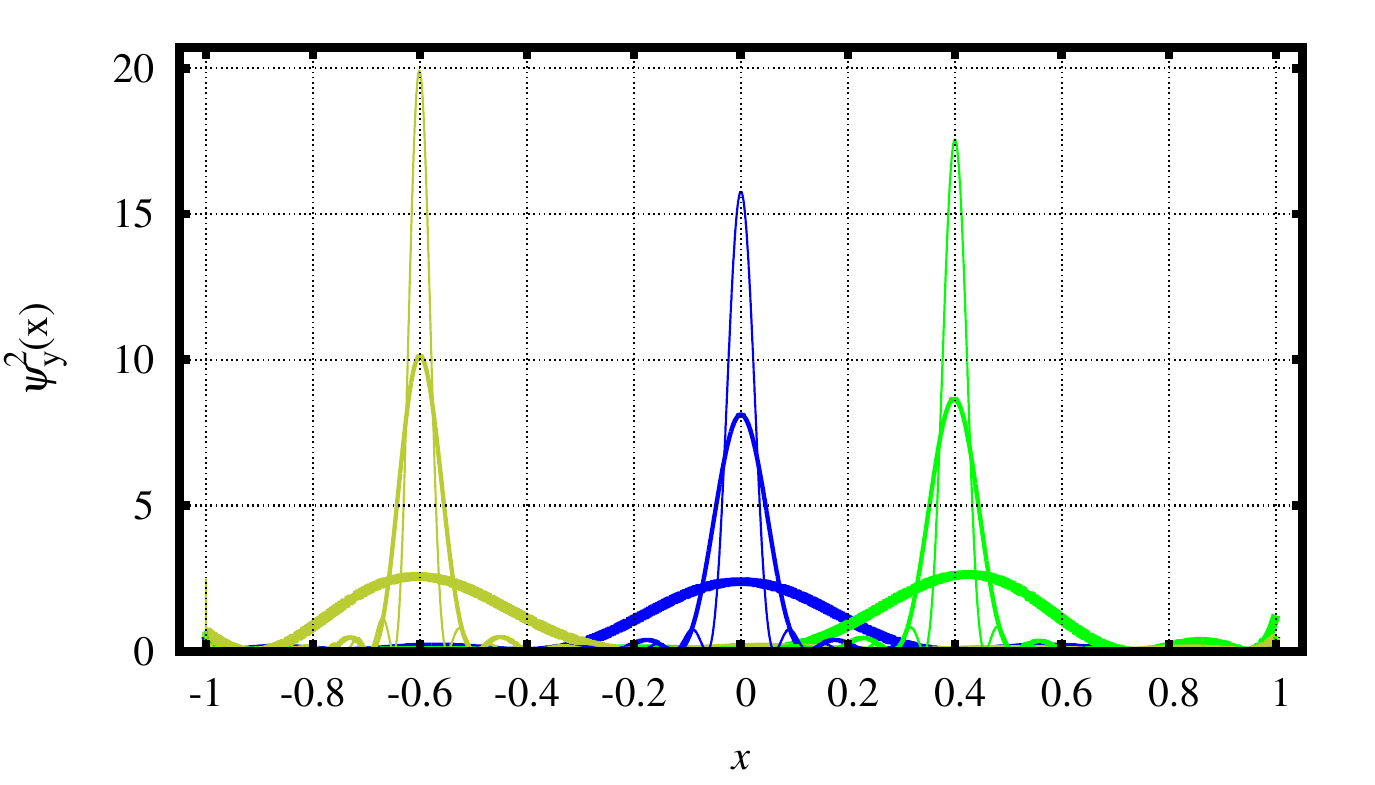}
  \caption{\label{Psi0}
  A simple demonstration of (\ref{psiYlocalized})
  localized states $\psi^2_{\mathbf{y}}(\mathbf{x})$
  for the measure $\Braket{g}=\int_{-1}^{1}g(x)dx$
  and the basis $\mathbf{x}$ constructed from 1D variable $x\in [-1:1]$
  as  $x_k=x^k$.
  The results for the states localized at $y= \{-0.6, 0, 0.4\}$ are presented
  as olive, blue, and green lines respectively.
  Basis dimension $n$ is chosen as $\{7, 25, 50\}$
  for thick, middle, and thin lines respectively.
  }
\end{figure}

\begin{figure}[t]
 \includegraphics[width=5.2cm]{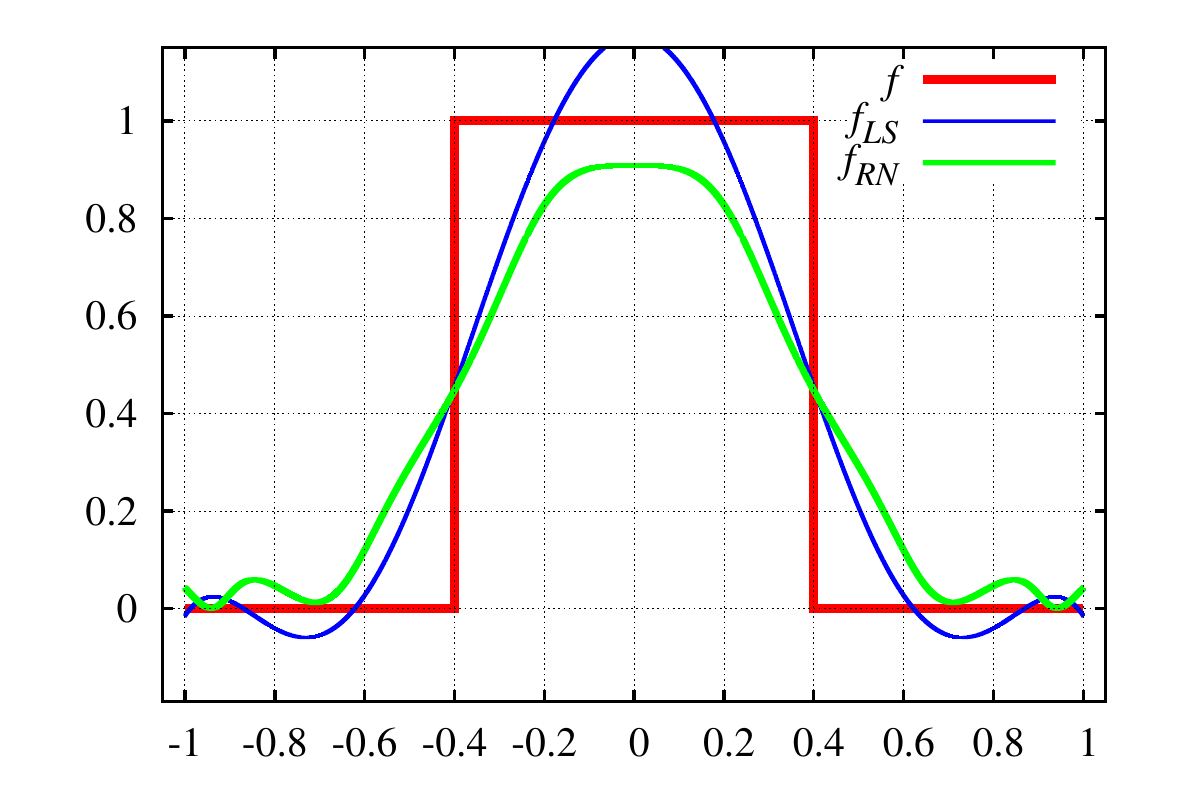}
 \includegraphics[width=5.2cm]{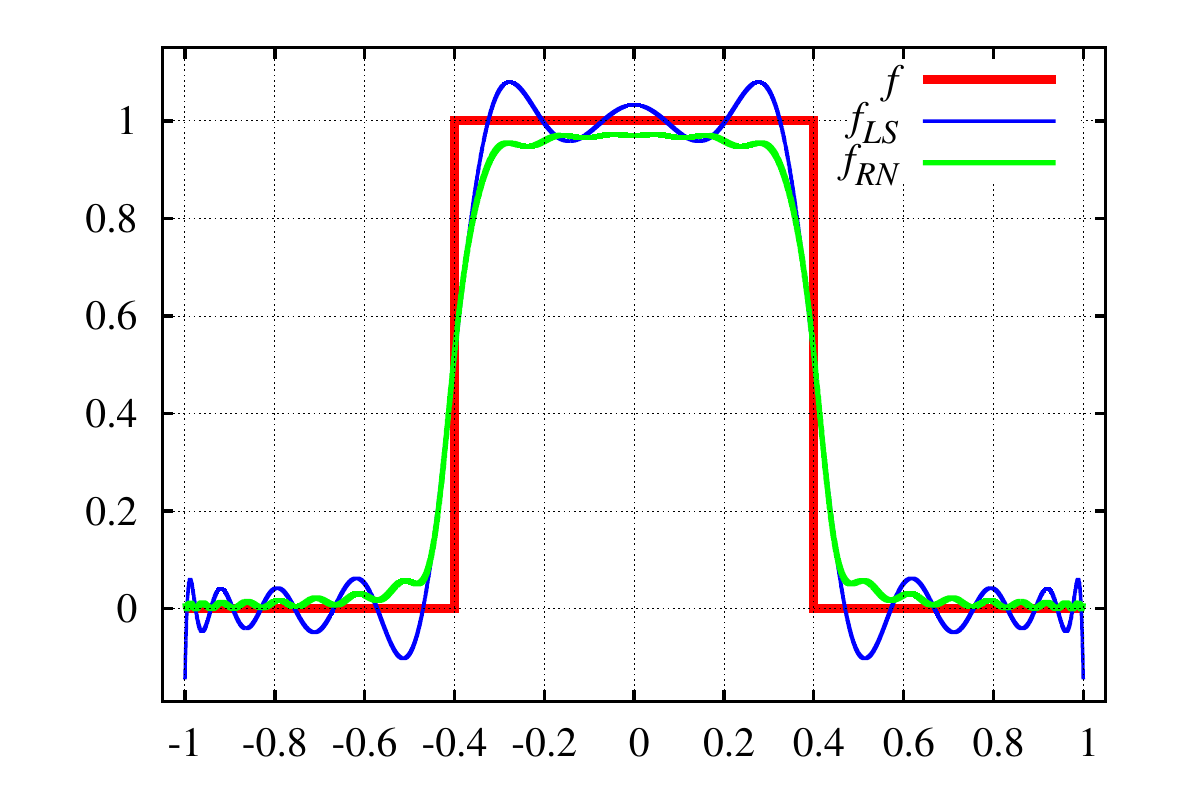}
 \includegraphics[width=5.2cm]{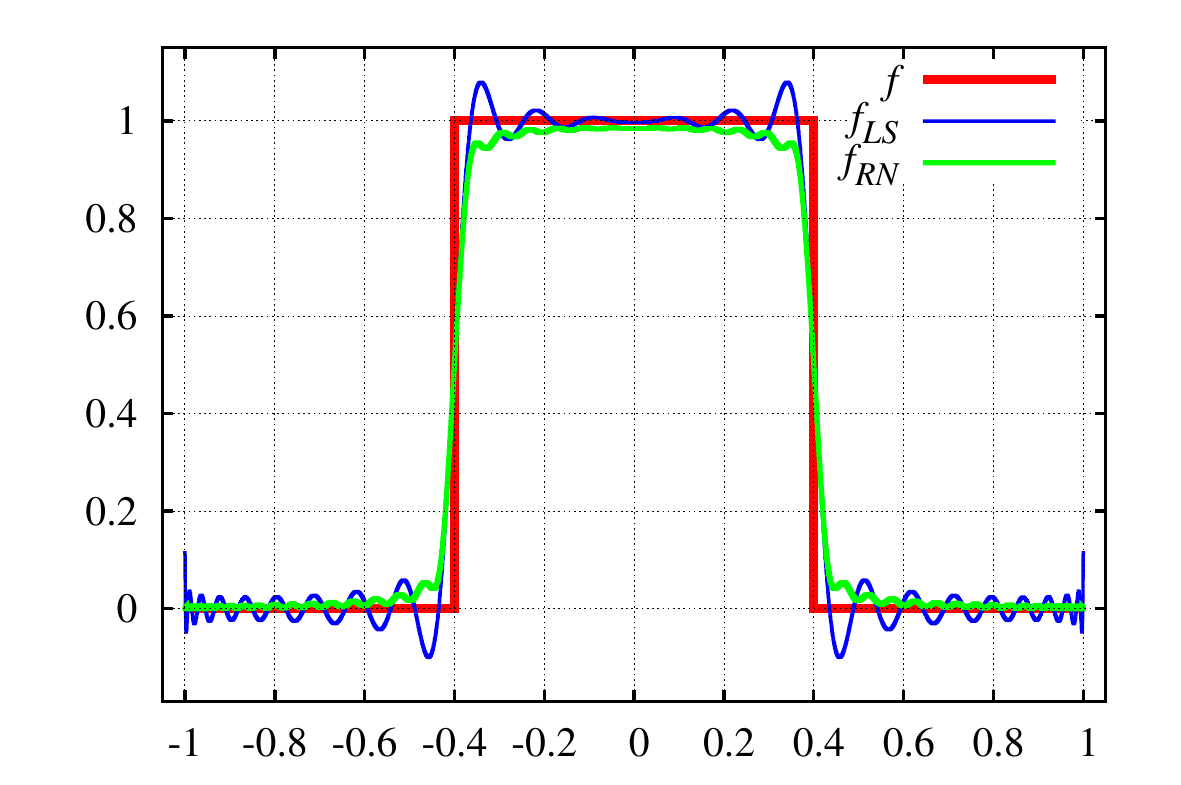}
  \caption{\label{RNapproximationSquareWave}
  A demonstration of a square wave interpolation (red) by
  least squares (blue, Eq. (\ref{fxapproxLS}))
  and
  Radon--Nikodym (green, Eq. (\ref{RNfsolutionX}))
  with the measure $\Braket{g}=\int_{-1}^{1}g(x)dx$
  with $x\in [-1:1]$ for $n=\{7, 25, 50\}$ in the pictures:
  left, middle, and right respectively.
  }
\end{figure}

A simple demonstration of localized states is presented
in Fig. \ref{Psi0}. For a simple chart a multi-dimensional
vector $\mathbf{x}$ is constructed from
1D variable $x\in [-1:1]$
as $x_k=x^k$. The measure $\Braket{\cdot}$
is taken as $\Braket{g}=\int_{-1}^{1}g(x)dx$.
Then $\psi^2_{\mathbf{y}}(\mathbf{x})$
can be considered
as a function of scalar $x$ and $y$ as $\mathbf{x}$ and $\mathbf{y}$ vectors
are calculated from the powers of $x$ and $y$.
In Fig. \ref{Psi0}  we present
$\psi^2_{-0.6}(x)$,
$\psi^2_{0}(x)$, and
$\psi^2_{0.4}(x)$.
As expected the $\psi^2_{y}(x)$ density is localized near $x=y$;
the localization becomes stronger with 
$n$ increase.
This chart demonstrates the main concept
behind Radon--Nikodym type of interpolation
which is a two--step process: on the first step a localized
state $\psi^2_{\mathbf{y}}(\mathbf{x})$ is built and on the second step
the value of an observable $f$ is evaluated at $\mathbf{y}$ by
averaging it with the weight obtained on the first step:
$f(\mathbf{y})\approx\Braket{\psi^2_{\mathbf{y}}f}\big/\Braket{\psi^2_{\mathbf{y}}}$.
A trivial example of a square wave interpolation using least squares and Radon--Nikodym
is presented in Fig. \ref{RNapproximationSquareWave}.
We see that Radon--Nikodym preserves the bounds of $f$ and has near interval edge
oscillations very much suppressed
because an interpolation of $f$ at $y$
is obtained by
averaging $f$ with always positive
weight $d\mu=\psi^2_{y}(x)dx$.

\subsection{\label{PureJointDistributionModel}Pure Joint Distribution Model}
In the section above we considered a simple problem of recovering
$\mathbf{f}$ from $\mathbf{x}$ given sampled data (\ref{mlproblemVector}).
The least squares and Radon--Nikodym estimators (\ref{fxapproxLS}) and (\ref{RNfsolutionX}) were obtained.
They are using individual components
of vector $\mathbf{f}$ as separate class labels;
vector class label makes the study much more difficult than a scalar one.
For further development we need,
for attributes $\mathbf{x}$ and class label $\mathbf{f}$ of vector type,
to have
estimators of joint distribution
$P(\mathbf{x},\mathbf{f})$ probability and corresponding to it coverage.

There are several possible approaches to unify $\mathbf{x}$ and $\mathbf{f}$.
In \cite{marx2019tractable} the authors introduced a new vector $\mathbf{z}$
of the dimension $n+m$
\begin{align}
\mathbf{z}&=(x_0,x_1,\dots,x_k,\dots,x_{n-1},f_0,f_1,\dots,f_j,\dots,f_{m-1})
\label{zDef}
\end{align}
and constructed Christoffel function from it
(this requires all
$\Braket{x_kx_{k^{\prime}}}$,
 $\Braket{f_jf_{j^{\prime}}}$, and $\Braket{x_kf_j}$ moments).
Maximizing Christoffel function
on $\mathbf{f}$ given $\mathbf{x}$ exhibits very promising
results. However, a difficulty with 
cross-terms arise\cite{malyshkin2019radonnikodym}
both in data initial regularization and in interpretation of the final result.

To deal with vector class label $\mathbf{f}$
and, for further generalization of Section \ref{KnowledgeGeneralizationOperator}
below, we will use $\mathbf{f}$-localized states.
For sampled $\mathbf{f}$ data, possibly 
\hyperref[aboutProductAttributes]{producted} to some order, construct
Gram matrix in $\mathbf{f}$-space (\ref{GramF}) and,
the same as in (\ref{psiYlocalized}),
build a localized state $\psi_{\mathbf{g}}(\mathbf{f})$:
\begin{align}
       \psi_{\mathbf{g}}(\mathbf{f})&=
           \frac{\sum\limits_{j,j^{\prime}=0}^{m-1}g_jG^{\mathbf{f};\,-1}_{jj^{\prime}}f_{j^{\prime}}}
           {\sqrt{\sum\limits_{j,j^{\prime}=0}^{m-1}g_jG^{\mathbf{f};\,-1}_{jj^{\prime}}g_{j^{\prime}}}}             
       =\frac{\sum\limits_{j=0}^{m-1}\psi^{[i]}(\mathbf{g})\psi^{[j]}(\mathbf{f})}
           {\sqrt{\sum\limits_{j=0}^{m-1}\left[\psi^{[j]}(\mathbf{g})\right]^2}}
  \label{psiGlocalized}
\end{align}
For each observation $l=1\dots M$ consider
(\ref{probgypkApp}) projection of $\mathbf{x}^{(l)}$--localized state
(\ref{psiYlocalized}) to $\mathbf{f}^{(l)}$--localized state
(\ref{psiGlocalized})
then sum it over the entire $l=1\dots M$ sample
to obtain the number of covered observations
${\cal F}^{JDG}$ (note: there is a ``projective'' factor $\Braket{f_j x_{k} }$
in the expression)
\begin{align}
  \Braket{\psi_{\mathbf{g}}|\psi_{\mathbf{y}}}^2&=
  \frac{
 \left|
    \sum\limits_{k,k^{\prime}=0}^{n-1}\sum\limits_{j,j^{\prime}=0}^{m-1}
g_{j^{\prime}}  G^{\mathbf{f};\,-1}_{j^{\prime}j} \Braket{f_j x_{k} }
G^{\mathbf{x};\,-1}_{kk^{\prime}}y_{k^{\prime}}
    \right|^2
  }
       {
         \sum\limits_{j,j^{\prime}=0}^{m-1} g_{j}G^{\mathbf{f};\,-1}_{jj^{\prime}}g_{j^{\prime}}
         \sum\limits_{k,k^{\prime}=0}^{n-1} y_{k}G^{\mathbf{x};\,-1}_{kk^{\prime}}y_{k^{\prime}}
       } \label{probgypkApp} \\
\Braket{1}\ge{\cal F}^{JDG}&=\sum\limits_{l=1}^{M}\Braket{\psi_{\mathbf{f}^{(l)}}|\psi_{\mathbf{x}^{(l)}}}^2
\omega^{(l)}
\label{allProj}
\end{align}
If $\mathbf{x}$ and $\mathbf{f}$ form the same vector space
then ${\cal F}^{JDG}=\Braket{1}$. Otherwise, for example when
$\mathbf{x}$ contains the entire $\mathbf{f}$ plus one more completely random attribute,
${\cal F}^{JDG}<\Braket{1}$.
Since (\ref{probgypkApp}) has normalizing terms containing
$G^{\mathbf{f};\,-1}_{jj^{\prime}}$ and $G^{\mathbf{x};\,-1}_{kk^{\prime}}$
matrices
in the denominator,
to calculate (\ref{allProj}) the secondary sampling technique\cite{MalMuseScalp}
is required. The Gram matrices
$G^{\mathbf{x}}_{kk^{\prime}}$ (\ref{GramX})
and 
$G^{\mathbf{f}}_{jj^{\prime}}$ (\ref{GramF})
are calculated first then
the projection (\ref{probgypkApp}) is calculated
for every observation $l=1\dots M$ and used in (\ref{allProj})
as it were plain observed at observation $l$.
Technically this means we need to calculate
the moments of $\mathbf{x}$- and $\mathbf{f}$- Christoffel functions product:
$\Braket{x_{k}f_{j}|K^{(\mathbf{x})}K^{(\mathbf{f})}|x_{k^{\prime}}f_{j^{\prime}}}$
(\ref{ChristoffelfunctionsProductMoments}).

The $\Braket{\psi_{\mathbf{g}}|\psi_{\mathbf{y}}}^2$
can be viewed as joint distribution of $\mathbf{f}$ and $\mathbf{x}$.
For a given $\mathbf{x}$ the probabilities of various $\mathbf{f}$
can be estimated as
\begin{align}
P(\mathbf{f},\mathbf{x})&\approx\Braket{\psi_{\mathbf{f}}|\psi_{\mathbf{x}}}^2
\label{ProbEstimation}
\end{align}
The (\ref{ProbEstimation}) estimates the probability
of possible outcome $\mathbf{f}$ given some fixed value of $\mathbf{x}$;
the estimation is based on (attributes, class label) pairs observed
in the training sample.
A typical step from here is to find a subspace of $\mathbf{x}$
providing the best prediction of $\mathbf{f}$,
optimal clustering \cite{malyshkin2019radonnikodym}
is a typical approach in this direction.
However, we want to go beyond 
``joint distribution generalization'',
beyond finding a subspace of $\mathbf{x}$
providing the best prediction of $\mathbf{f}$
in terms of the probability $P(\mathbf{f},\mathbf{x})$
estimated on \textsl{training sample}. We need a more powerful
generalization method,
possibly applicable to not yet observed
values of $\mathbf{x}$ and $\mathbf{f}$.

\section{\label{KnowledgeGeneralizationOperator}On Knowledge Generalizing Operator}
In Section \ref{PureJointDistributionModel} above
we transformed original
$\mathbf{x}^{(l)}\to\mathbf{f}^{(l)}$
data sample (\ref{mlproblemVector})
to a sequence of $\mathbf{x}^{(l)}$- and $\mathbf{f}^{(l)}$-
localized states according to (\ref{psiYlocalized}) and (\ref{psiGlocalized}):
\begin{align}
\psi_{\mathbf{x}^{(l)}}&\to
\psi_{\mathbf{f}^{(l)}}
  & \text{weight $\omega^{(l)}$}  \label{mlproblemVectorPSI}
\end{align}
As $\psi_{\mathbf{x}}$ / $\psi_{\mathbf{f}}$
are defined by $n$ / $m$ coefficients before $x_k$ / $f_j$
the (\ref{mlproblemVectorPSI}) is nothing more than
a transform of the original data (\ref{mlproblemVector}).
This is not a regular linear transform of
$\mathbf{x}$ / $\mathbf{f}$ bases,
this is a linear transform with 
$G^{\mathbf{x};\,-1}_{kk^{\prime}}$ / $G^{\mathbf{f};\,-1}_{jj^{\prime}}$ matrices
\textsl{followed by} normalization to $1$ with Christoffel function
as in (\ref{psiYlocalized}) / (\ref{psiGlocalized}).

The purpose of this transform is to obtain the states
we can project to each other or to some other basis;
for example:
$\Ket{\psi_{\mathbf{x}^{(l)}}}=\sum_{k=0}^{n-1}\Ket{\psi^{[k]}}\Braket{\psi^{[k]}|\psi_{\mathbf{x}^{(l)}}}$
with $\Ket{\psi^{[k]}}$ being an orthogonal full basis in $\mathbf{x}$-space,
$1=\Braket{\psi_{\mathbf{x}^{(l)}}|\psi_{\mathbf{x}^{(l)}}}^2$,
$1=\Braket{\psi_{\mathbf{f}^{(l)}}|\psi_{\mathbf{f}^{(l)}}}^2$,
$1\ge\Braket{\psi_{\mathbf{x}^{(l)}}|\psi_{\mathbf{f}^{(l)}}}^2$,
etc.
The main result of Section \ref{PureJointDistributionModel}
was to obtain joint $(\mathbf{x},\mathbf{f})$ distribution (\ref{ProbEstimation})
and then trying to generalize from it.

Consider a different form of generalization. Let $\Ket{\psi_{\mathbf{x}^{(l)}}}$,
before being used in calculation of joint distribution,
is transformed by a unitary operator $\|\mathcal{U}\|$:
\begin{align}
\mathcal{F}&=\sum\limits_{l=1}^{M}\Braket{\psi_{\mathbf{f}^{(l)}}|\mathcal{U}|\psi_{\mathbf{x}^{(l)}}}^2
\omega^{(l)}
\label{allProjU}
\end{align}
Contrary to (\ref{ProbEstimation})
this expression
is transforming $\psi_{\mathbf{x}^{(l)}}(\mathbf{x})$
to some other function in $\mathbf{x}$-space  $\Ket{\psi(\mathbf{x})}=
\Ket{\mathcal{U}|\psi_{\mathbf{x}^{(l)}}(\mathbf{x})}$
and only then projecting the result
to
actual realization $\psi_{\mathbf{f}^{(l)}}(\mathbf{f})$
in $\mathbf{f}$-space.
In some sense the $\|\mathcal{U}\|$ can be viewed
as a
\href{https://en.wikipedia.org/wiki/S-matrix#Definition}{Scattering Amplitude Matrix}, as it relates the
\emph{IN} state $\Ket{\psi_{\mathbf{x}^{(l)}}}$
with the \emph{OUT} state $\Ket{\psi_{\mathbf{f}^{(l)}}}$.
All the information about what combinations
of attributes $x_k$ to be used for prediction
now contains in operator $\|\mathcal{U}\|$. It is called
\textsl{Knowledge Generalizing Operator}.
The operator is unitary (to preserve normalizing)
$1=\Braket{\psi|\mathcal{U}^{\dagger}|\mathcal{U}|\psi}$.
\begin{align}
\mathcal{U}^{\dagger}\mathcal{U}&=\mathds{1} \label{constRaintUnitarity}
\end{align}
In our model the knowledge is represented in the form
of a unitary operator. This is a very common form in physics:
the dynamics 
in
classical mechanics, electrodynamics, quantum mechanics
can be represented as a sequence of infinitesimal unitary transformations
determined by Hamiltonian (or Lagrangian) of the system.
The (\ref{allProjU}) is an inverse problem:
given (\ref{mlproblemVector}) data find unitary
operator $\|\mathcal{U}\|$ maximizing (\ref{allProjU}) coverage
subject to (\ref{constRaintUnitarity}) constraints.
Whereas the (\ref{allProjU}) is of fourth order in $\Ket{\psi}$,
it is of second order in $\|\mathcal{U}\|$.
The constraints (\ref{constRaintUnitarity})
is also of second order in $\|\mathcal{U}\|$.
Thus the problem of finding
the \textsl{Knowledge Generalizing Operator}
is a variant of
quadratically constrained quadratic program\cite{shor2013nondifferentiable} (\href{https://en.wikipedia.org/wiki/Quadratically_constrained_quadratic_program}{QCQP}).

Before we go further, let us consider a simplified version
of (\ref{allProjU}) to obtain $\mathcal{F}$ upper limit
for ``projective'' interpretation of operator $\|\mathcal{U}\|$.
Consider the problem of finding (in $\mathbf{x}$-space)
orthogonal basis $\phi^{[i]}$,
 a subset of full basis $D\le n$,
that maximizes $\mathcal{F}^{TOT}$:
\begin{align}
 \phi^{[i]}&=\sum_{k=0}^{n-1}\alpha^{\phi;[i]}_kx_k & i=0\dots D-1 ; \enspace D\le n \label{phiiD}\\
 \delta_{ii^{\prime}}&=\Braket{\phi^{[i]}|\phi^{[i^{\prime}]}}=
 \sum_{k,k^{\prime}=0}^{n-1}\alpha^{\phi;[i]}_k G^{\mathbf{x}}_{kk^{\prime}} \alpha^{\phi;[i^{\prime}]}_{k^{\prime}} \label{pisiDnorm}\\
  \mathcal{F}^{TOT}&=
\sum\limits_{l=1}^{M}\sum\limits_{i=0}^{D-1}\Braket{\psi_{\mathbf{f}^{(l)}}|\phi^{[i]}}^2
\omega^{(l)}
 \label{FtotalFullBasis}
\end{align}
Substituting (\ref{phiiD}) to (\ref{psiGlocalized}) obtain:
\begin{align}
K^{(\mathbf{f})}(\mathbf{g})&=\frac{1}{
\sum\limits_{j,j^{\prime}=0}^{m-1} g_{j}G^{\mathbf{f};\,-1}_{jj^{\prime}}g_{j^{\prime}}
}
\label{ChristoffelF}\\
\Braket{f_{t}|K^{(\mathbf{f})}|f_{s}}&=
\sum\limits_{l=1}^{M}
\frac{f^{(l)}_{t}f^{(l)}_{s}}
{
\sum\limits_{j,j^{\prime}=0}^{m-1} f^{(l)}_{j}G^{\mathbf{f};\,-1}_{jj^{\prime}}f^{(l)}_{j^{\prime}}
}\omega^{(l)} \label{ChristoffelFunMoments}\\
K^{(\mathbf{f\to x})}_{ik}&=
\sum\limits_{k^{\prime},t^{\prime},s^{\prime},j^{\prime}=0}^{m-1}
\Braket{x_i f_{k^{\prime}}}
      G^{\mathbf{f};\,-1}_{k^{\prime}t^{\prime}}
      \Braket{f_{t^{\prime}}|K^{(\mathbf{f})}|f_{s^{\prime}}}
      G^{\mathbf{f};\,-1}_{s^{\prime}j^{\prime}}
      \Braket{x_k f_{j^{\prime}}}
       &i,k=0\dots n-1 \label{Kftoxsum}
\end{align}
The (\ref{ChristoffelFunMoments}) is $\mathbf{f}$--Christoffel function
(\ref{ChristoffelF})
moments\footnote{
One can also consider
$\Braket{\frac{\partial R}{\partial f_t}\frac{\partial R}{\partial f_s}}$
with $R(\mathbf{f})=1/\sqrt{K(\mathbf{f})}=
\sqrt{\sum\limits_{j,j^{\prime}=0}^{m-1} f_{j}G^{\mathbf{f};\,-1}_{jj^{\prime}}f_{j^{\prime}}}$.
}.
The (\ref{Kftoxsum}) allows to present (\ref{FtotalFullBasis})
in the form:
\begin{align}
 \mathcal{F}^{TOT}&=
 \sum\limits_{i=0}^{D-1} \sum_{k,k^{\prime}=0}^{n-1}\alpha^{\phi;[i]}_k
 K^{(\mathbf{f\to x})}_{kk^{\prime}}\alpha^{\phi;[i]}_{k^{\prime}}
 \label{FtotExpansion}
\end{align}
From which we can spectrally expand the $\mathcal{F}^{TOT}$
by solving a
\href{https://en.wikipedia.org/wiki/Eigendecomposition_of_a_matrix\#Generalized_eigenvalue_problem}{generalized eigenvalue problem}
with the matrices $K^{(\mathbf{f\to x})}_{kk^{\prime}}$
and $G^{\mathbf{x}}_{kk^{\prime}}$ in left- and right- hand sides:
\begin{align}
  \sum\limits_{k^{\prime}=0}^{n-1} K^{(\mathbf{f\to x})}_{kk^{\prime}}
  \alpha^{\phi;[i]}_{k^{\prime}} &=
  \lambda^{[i]} \sum\limits_{k^{\prime}=0}^{n-1} G^{\mathbf{x}}_{kk^{\prime}}
  \alpha^{\phi;[i]}_{k^{\prime}}
\label{GEVKftoXfxf} \\
\mathcal{F}^{TOT}&=\sum\limits_{i=0}^{D-1} \lambda^{[i]}
\label{FtotDecompose} 
\end{align}
The (\ref{FtotDecompose}) is a spectral decomposition of (\ref{FtotalFullBasis}),
it has at most $m$ non--zero eigenvalues (the rank of (\ref{Kftoxsum}) is $m$ or lower, we also assume $m\le n$).
If $\mathbf{f}$ belongs to a subspace of $\mathbf{x}$ then
the sum of these $m$ eigenvalues in (\ref{FtotDecompose})
is equal to $\Braket{1}$.
The (\ref{FtotDecompose}) takes all possible
vectors from $\mathbf{x}$-space and project them to
all $\Ket{\psi_{\mathbf{f}^{(l)}}}$ summing the coverage, this operation
does not make any inference, it estimates
the coverage (\ref{allProjU})
upper limit for any norm--preserving
projective transform\cite{gsmalyshkin2017comparative},
such as $\Ket{\mathcal{U}|\psi}$ (\ref{Uevolution}) or, more generally, (\ref{KrausOperator}).
The estimation can be obtained 
from $K^{(\mathbf{f\to x})}_{kk^{\prime}}$
and $G^{\mathbf{x}}_{kk^{\prime}}$ matrices even without solving
the eigenvalue problem (\ref{GEVKftoXfxf}):
\begin{align}
\mathcal{F}^{TOT}&=\sum\limits_{k,k^{\prime}=0}^{n-1}
K^{(\mathbf{f\to x})}_{kk^{\prime}}
G^{\mathbf{x};\,-1}_{k^{\prime}k}
\label{FtotWithoutEigenproblem}
\end{align}
For calculation see
\texttt{\seqsplit{com/polytechnik/utils/KGOSolutionVectorXVectorF.java:FTOT}}
which is used in unit tests.

A simpler approach to construct
contributing to coverage subspace  $\Ket{\phi^{[i]}}$
is to notice that in (\ref{allProjU})
there are scalar products $\Braket{f_j x_{k} }$ of the
vectors from 
$\mathbf{x}$ and $\mathbf{f}$ spaces.
Thus we can project the $\mathbf{f}$-space to $\mathbf{x}$-space;
to split $\mathbf{x}$ into two subspaces:
$\Ket{\phi^{P;[j]}}$ ``projected'' (of the dimension $D\le m$)
and $\Ket{\phi^{O;[k]}}$ ``orthogonal'' to $\mathbf{f}$ (of the dimension $n-D$),
all vectors from the second one have zero scalar product
with a state in $\mathbf{f}$-space $\Braket{\phi^{O;[k]}|\psi_{\mathbf{f}}}=0$,
thus the $\Ket{\phi^{O;[k]}}$ does not contribute to coverage (\ref{allProjU}).
For this reason it is sufficient to consider operator $\|\mathcal{U}\|$
to have the dimension $D\times n$ converting
a vector from  $\mathbf{x}$-space to $\Ket{\phi^{P;[j]}}$,
i.e. to use $\Ket{\phi^{[i]}}=\Ket{\phi^{P;[i]}}$
as contributing subspace,
see
\texttt{\seqsplit{com/polytechnik/utils/TestKGO.java:orthogonalizeU}}
for an implementation.

Let us define operator $\|\mathcal{U}\|$ to be a matrix
(in this paper $u_{sk}$ is considered to be a real matrix,
a generalization to a complex matrix is straightforward)
of $D\times n$, $D\le m$,
such that:
\begin{align}
\Ket{\mathcal{U}|x_k}&=\sum\limits_{s=0}^{D-1} \Ket{\phi^{[s]}} u_{sk}
\label{UdefX}
\end{align}
Then (note: there is a ``projective'' factor
$\Braket{f_j x_{k} }$
in the expression, from  $\Braket{f_j \phi^{[s]} }$)
\begin{align}
  \Braket{\psi_{\mathbf{g}}|\mathcal{U}|\psi_{\mathbf{y}}}^2
&=
  \frac{
 \left|
    \sum\limits_{k,k^{\prime}=0}^{n-1}\sum\limits_{j,j^{\prime}=0}^{m-1}
    \sum\limits_{s=0}^{D-1}
g_{j^{\prime}}  G^{\mathbf{f};\,-1}_{j^{\prime}j} \Braket{f_j \phi^{[s]} }
u_{sk}
G^{\mathbf{x};\,-1}_{kk^{\prime}}y_{k^{\prime}}
    \right|^2
  }
       {
         \sum\limits_{j,j^{\prime}=0}^{m-1} g_{j}G^{\mathbf{f};\,-1}_{jj^{\prime}}g_{j^{\prime}}
         \sum\limits_{k,k^{\prime}=0}^{n-1}
         y_{k} G^{\mathbf{x};\,-1}_{kk^{\prime}}y_{k^{\prime}}
       } \label{probgUypkApp} \\
\Braket{\phi^{[s]}|\phi^{[q]}}&= \sum\limits_{k,k^{\prime} =0}^{n-1}u_{sk}
\Braket{x_k|x_{k^{\prime}}}
u_{qk^{\prime}}& s,q=0\dots D-1
    \label{optimmatrixConstraintAppendixNU}
\end{align}
The (\ref{probgUypkApp}) is actually (\ref{probgypkApp})
but $\Ket{\psi_{\mathbf{y}}}$
is replaced by
$\Ket{\mathcal{U}|\psi_{\mathbf{y}}}$.
This is the central concept
of knowledge generalizing operator:
the state the inference is based on $\Ket{\psi_{\mathbf{y}}}$
is transformed by the operator $\|\mathcal{U}\|$
before coupling with the state $\Ket{\psi_{\mathbf{f}}}$ we are looking an inference to.
Partial unitarity constraint (\ref{optimmatrixConstraintAppendixNU})
corresponds to the fact that 
only subspace
of the dimension $D\le m$ can possibly contribute to the coverage (\ref{allProjU}).
When only a subspace of $\mathbf{x}$ contributes to
(\ref{allProjUKxf})
the problem to find a unitary matrix $u_{jk}$ becomes highly degenerative.
While the algorithm described in the Appendix \ref{numAlgorithm}
below
works well with such a degenerative problem,
it is beneficial for both: computational complexity and
simplicity of result's interpretation
to make the problem less degenerative.
Consider a  $\Ket{\phi^{[i]}}$ $i=0\dots D-1$
 subspace
of the dimension $D\le m$.
Let us split
considered above
unitary operator $\|\mathcal{U}\|$
into $\|\mathcal{U}\|=\|\mathcal{U}^P\|+\|\mathcal{U}^O\|$
such that $\|\mathcal{U}^P\|$ transforms any $\mathbf{x}$-vector to $\Ket{\phi^{[i]}}$ subspace,
and $\|\mathcal{U}^O\|$ transforms any $\mathbf{x}$-vector
to a subspace orthogonal to $\Ket{\phi^{[i]}}$ (this split is most easy to perform
if to convert original
$\Ket{x_k}$ space into direct sum of $\Ket{\phi^{[i]}}$ and orthogonal to $\Ket{\phi^{[i]}}$ subspaces).
Then, because $0=\Braket{\psi_{\mathbf{f}}|\mathcal{U}^O|\psi_{\mathbf{x}}}$
for any $\mathbf{f}$ and $\mathbf{x}$,
optimization result of $\mathcal{F}$ does not depend on $\|\mathcal{U}^O\|$,
thus it is sufficient to find an operator $\|\mathcal{U}^P\|$
of the dimension $D\times n$ subject to (\ref{optimmatrixConstraintAppendixNU}) constraint.

To calculate (\ref{allProjU}) it is convenient to introduce the moments
of Christoffel functions product:
\begin{align}
\Braket{x_{k}f_{j}|K^{(\mathbf{x})}K^{(\mathbf{f})}|x_{k^{\prime}}f_{j^{\prime}}}&=
\sum\limits_{l=0}^{M}\omega^{(l)}
\frac{x^{(l)}_{k}x^{(l)}_{k^{\prime}}}
{
         \sum\limits_{q,q^{\prime}=0}^{n-1}
         x^{(l)}_{q} G^{\mathbf{x};\,-1}_{qq^{\prime}}x^{(l)}_{q^{\prime}}
}
\cdot
\frac{f^{(l)}_{j}f^{(l)}_{j^{\prime}}}
{
\sum\limits_{s,s^{\prime}=0}^{m-1} f^{(l)}_{s}G^{\mathbf{f};\,-1}_{ss^{\prime}}f^{(l)}_{s^{\prime}}
}
\label{ChristoffelfunctionsProductMoments}
\end{align}
to write $\mathcal{F}$ in the form
(note: there is a ``projective'' factor $\Braket{f_j x_{k} }$
in the expression, from  $\Braket{f_j \phi^{[s]} }$)
\begin{align}
S_{sk;s^{\prime}k^{\prime}}&=
\sum\limits_{j,j^{\prime},q,q^{\prime}=0}^{m-1}
\sum\limits_{t,t^{\prime}=0}^{n-1}
\Braket{x_{t}f_{j^{\prime}}|K^{(\mathbf{x})}K^{(\mathbf{f})}|x_{t^{\prime}}f_{q^{\prime}}}
G^{\mathbf{f};\,-1}_{j^{\prime}j} \Braket{f_j \phi^{[s]} } G^{\mathbf{x};\,-1}_{kt}
G^{\mathbf{f};\,-1}_{q^{\prime}q} \Braket{f_q \phi^{[s^{\prime}]} } G^{\mathbf{x};\,-1}_{k^{\prime}t^{\prime}}
\label{Smatrix}\\
\mathcal{F}&=\sum\limits_{s,s^{\prime}=0}^{D-1}\sum\limits_{k,k^{\prime}=0}^{n-1}
u_{sk}S_{sk;s^{\prime}k^{\prime}}u_{s^{\prime}k^{\prime}}
=\sum\limits_{l=1}^{M}\Braket{\psi_{\mathbf{f}^{(l)}}|\mathcal{U}|\psi_{\mathbf{x}^{(l)}}}^2
\omega^{(l)}
\xrightarrow[\mathcal{U}]{\quad }\max
\label{allProjUKxf}
\end{align}
The $\mathcal{F}$ is a quadratic function on $u_{sk}$;
the expression for $S_{sk;s^{\prime}k^{\prime}}$ can be greatly
simplified if $\mathbf{x}$- and $\mathbf{f}$- bases are
initially regularized (see \cite{malyshkin2019radonnikodym},
 ``\emph{Appendix A: Regularization Example}'').
 In an orthogonal basis Gram matrix is a unit matrix,
thus the $G^{\mathbf{x};\,-1}$ and $G^{\mathbf{f};\,-1}$ get removed in (\ref{Smatrix}).

\subsection{\label{KGOWithoutContributingSubspace}On Knowledge Generalizing Operator
With Different \emph{IN} and \emph{OUT} Spaces}
In the section
above we considered operator $\|\mathcal{U}\|$ as
$\mathbf{x}\to\mathbf{x}$ transform. In (\ref{allProjU}) the
$\Braket{\psi_{\mathbf{f}}|\mathcal{U}|\psi_{\mathbf{x}}}$
was understood as $\mathbf{x}\to\mathbf{x}$ transform
$\Ket{\mathcal{U}|\psi_{\mathbf{x}}}$
followed by projection of the result to
$\Ket{\psi_{\mathbf{f}}}$-space (\ref{probgUypkApp});
similar ``projective'' interpretation was used in
(\ref{probgypkApp}).
This interpretation of $\|\mathcal{U}\|$
lead us to ``contributing subspace'' $\Ket{\phi^{[s]}}$
(\ref{phiiD}) (which is a subspace of $\mathbf{x}$), equation
(\ref{probgUypkApp}) for $\Braket{\psi_{\mathbf{f}}|\mathcal{U}|\psi_{\mathbf{x}}}^2$
(it has $\Braket{f_j x_k }$ projective factors)
and (\ref{optimmatrixConstraintAppendixNU}) constraints
with the meaning of scalar product invariance.
Optimization problem (\ref{allProjUKxf})
for $u_{jk}$ matrix of the dimension $D\times n$
allows  to determine
partially unitary 
operator $\|\mathcal{U}\|$.
This operator has
both \emph{IN} and \emph{OUT} subspaces being a subspace of $\mathbf{x}$.

A natural generalization is to consider an operator $\|\mathcal{U}\|$
with \textsl{different} subspaces for \emph{IN} and \emph{OUT},
this way we can avoid any kind of ``projection''
what would greatly increase generalizing power of the approach.
Let us  consider $\mathbf{x}\to\mathbf{f}$
transform directly. Now $u_{jk}$ is a $m\times n$ matrix
transforming a vector from $\mathbf{x}$-space to $\mathbf{f}$-space
\begin{align}
f_j&=\sum\limits_{k=0}^{n-1}u_{jk}x_k
& j=0\dots m-1
\label{fProjXDifferently}
\end{align}
In a common ``projective'' paradigm the
(\ref{fProjXDifferently})
is multiplied by $x_{k^{\prime}}$, then 
after taking the average --- least squares (\ref{fxapproxLS})
are obtained. Now it is different --- we cannot take scalar products $\Braket{f_jx_k}$
as $\mathbf{f}$ and $\mathbf{x}$ belong to different Hilbert spaces.
We
multiply (\ref{fProjXDifferently})
by \textsl{itself} and take the average ---
obtain (\ref{optimmatrixConstraintAppendixNUAS}) constraint.
Substituting (\ref{fProjXDifferently})
to localized state (\ref{psiGlocalized}) obtain
\begin{align}
  \Braket{\psi_{\mathbf{g}}|\mathcal{U}|\psi_{\mathbf{y}}}^2
&=
  \frac{
 \left|
    \sum\limits_{k=0}^{n-1}\sum\limits_{j,s=0}^{m-1}
g_{j}  G^{\mathbf{f};\,-1}_{js}
u_{sk}
y_{k}
    \right|^2
  }
       {
         \sum\limits_{j,j^{\prime}=0}^{m-1} g_{j}G^{\mathbf{f};\,-1}_{jj^{\prime}}g_{j^{\prime}}
         \sum\limits_{k,k^{\prime}=0}^{n-1}
         y_{k} G^{\mathbf{x};\,-1}_{kk^{\prime}}y_{k^{\prime}}
       } \label{probgUypkAppAS} \\
\Braket{f_j|f_{j^{\prime}}}&= \sum\limits_{k,k^{\prime} =0}^{n-1}u_{jk}
\Braket{x_k|x_{k^{\prime}}}
u^*_{j^{\prime}k^{\prime}}& j,j^{\prime}=0\dots m-1
    \label{optimmatrixConstraintAppendixNUAS}
\end{align}
thus the optimization problem
does not contain any
``projective'' factors $\Braket{f_j x_{k} }$
\begin{align}
S_{sk;s^{\prime}k^{\prime}}&=
\sum\limits_{j,j^{\prime}=0}^{m-1}
\Braket{f_{j}x_{k}|K^{(\mathbf{x})}K^{(\mathbf{f})}|f_{j^{\prime}}x_{k^{\prime}}}
G^{\mathbf{f};\,-1}_{js}
G^{\mathbf{f};\,-1}_{j^{\prime}s^{\prime}}
\label{SmatrixAS}\\
\mathcal{F}&=\sum\limits_{s,s^{\prime}=0}^{m-1}\sum\limits_{k,k^{\prime}=0}^{n-1}
u_{sk}S_{sk;s^{\prime}k^{\prime}}u^*_{s^{\prime}k^{\prime}}
=\sum\limits_{l=1}^{M}\Braket{\psi_{\mathbf{f}^{(l)}}|\mathcal{U}|\psi_{\mathbf{x}^{(l)}}}^2
\omega^{(l)}
\xrightarrow[\mathcal{U}]{\quad }\max
\label{allProjUKxfAS}
\end{align}
This is \textsl{the equation}.
The $\Braket{\psi_{\mathbf{f}}|\mathcal{U}|\psi_{\mathbf{x}}}$
is interpreted as operator $\|\mathcal{U}\|$
relating the states from
two \textsl{different} Hilbert space,
a
type of 
\href{https://en.wikipedia.org/wiki/Quantum_channel#Heisenberg_picture}{memoryless quantum channel},
a map between two spaces of operators.
Every admissible transformation $u_{jk}$
must satisfy Gram matrix invariance condition (\ref{optimmatrixConstraintAppendixNUAS}).
This condition can be satisfied only for $m\le n$
since $\Braket{f_j|f_{j^{\prime}}}$ has the rank $m$
and the matrix in the right hand side has the rank not greater than $n$;
in case $m>n$ one can consider (\ref{xfTransform}) and obtain (\ref{xfprobgUypkAppAS})
\begin{align}
x_k&=\sum\limits_{j=0}^{m-1}u_{kj}f_j
\label{xfTransform}\\
\Braket{x_k|x_{k^{\prime}}}&= \sum\limits_{j,j^{\prime} =0}^{m-1}u_{kj}
\Braket{f_j|f_{j^{\prime}}}
u_{k^{\prime}j^{\prime}}^*& k,k^{\prime}=0\dots n-1
    \label{xfoptimmatrixConstraintAppendixNUAS}\\
  \Braket{\psi_{\mathbf{g}}|\mathcal{U}|\psi_{\mathbf{y}}}^2
&=
  \frac{
 \left|
    \sum\limits_{j=0}^{m-1}\sum\limits_{k,q=0}^{n-1}
    y_{k}
 G^{\mathbf{x};\,-1}_{kq}
u_{qj}
g_{j}
    \right|^2
  }
       {
         \sum\limits_{j,j^{\prime}=0}^{m-1} g_{j}G^{\mathbf{f};\,-1}_{jj^{\prime}}g_{j^{\prime}}
         \sum\limits_{k,k^{\prime}=0}^{n-1}
         y_{k} G^{\mathbf{x};\,-1}_{kk^{\prime}}y_{k^{\prime}}
       } \label{xfprobgUypkAppAS} 
\end{align}
Thus it is sufficient just to swap $\mathbf{x}$ and $\mathbf{f}$
in numerical calculations.
When working in orthogonal bases $\delta_{kk^{\prime}}=\Braket{x_k|x_{k^{\prime}}}$
and $\delta_{jj^{\prime}}=\Braket{f_j|f_{j^{\prime}}}$
the matrix
elements of $S_{sk;s^{\prime}k^{\prime}}$  are (\ref{ChristoffelfunctionsProductMoments}).
Also see Appendix \ref{PAdjusted} below for possible
adjustment of probability normalizing.

Mapping an operator $A$ between $\mathbf{x}$- and  $\mathbf{f}$-
spaces is the same  transformation
$A^{\mathbf{f}}_{jj^{\prime}}=
\sum_{k,k^{\prime} =0}^{n-1}u_{jk}
A^{\mathbf{x}}_{kk^{\prime}}
u^*_{j^{\prime}k^{\prime}}$
as for Gram matrix (\ref{optimmatrixConstraintAppendixNUAS}).
The optimization problem  (\ref{allProjUKxfAS})
has the meaning of finding a quantum channel
conveying the highest possible probability
from  $\mathbf{x}$--space to $\mathbf{f}$--space.
A remarkable feature of this problem
is that it does not  contain any $\Braket{f_jx_k}$
averages! All the $\mathbf{x}\to\mathbf{f}$ inference
(communication between two ends of quantum channel)
now contains \textsl{only} in operator $\|\mathcal{U}\|$ ---
a matrix $u_{jk}$
of the dimension $m\times n$ to find
from optimization problem (\ref{allProjUKxfAS}).
This is an important new result.
In \cite{malyshkin2019radonnikodym} coverage optimization problem was
always formulated with some kind of  $\mathbf{x}\to\mathbf{f}$ projection;
if a model has $\Braket{f_j x_{k} }$ terms  -- it is
of ``projective'' type such as (\ref{probgypkApp}), (\ref{probgUypkApp})
or (\ref{Smatrix}) above.
The (\ref{probgUypkAppAS})
and (\ref{probgUypkAppASimportantOnly})
probabilities do not have $\Braket{f_j x_{k} }$ terms;
operator $\|\mathcal{U}\|$ 
directly (\ref{fProjXDifferently}) relates
$\mathbf{x}$- and $\mathbf{f}$- spaces
subject to 
(\ref{optimmatrixConstraintAppendixNUAS})
scalar product invariance;
it is the only link between \emph{IN} and \emph{OUT} spaces.
Familiar least squares expansion (\ref{fxapproxLS}) satisfies
the required constraints (\ref{optimmatrixConstraintAppendixNUAS})
\begin{align}
\Braket{f_j|f_{j^{\prime}}}&=
\sum\limits_{k,k^{\prime} =0}^{n-1}
\Braket{f_jx_k}G^{\mathbf{x};\,-1}_{kk^{\prime}}
\Braket{x_{k^{\prime}}f_{j^{\prime}}}
\label{leastsquares}
\end{align}
only when $\mathbf{f}$ is a subspace of $\mathbf{x}$;
Proof:
select some orthogonal bases such as $\delta_{kk^{\prime}}=\Braket{x_k|x_{k^{\prime}}}$
and $\delta_{jj^{\prime}}=\Braket{f_j|f_{j^{\prime}}}$,
obtain $1=\sum_{k=0}^{n-1}\Braket{f_jx_k}^2$,
i.e. only when $\mathbf{x}\to\mathbf{f}$ least squares mapping is exact.
Note that one can always apply Appendix (\ref{numAlgorithmApproximate})
method of singular values adjustment to obtain a partially unitary
transform from the least squares
or any other mapping that initially does not satisfy the partial
unitarity constraints (\ref{optimmatrixConstraintAppendixNUAS}).

The  optimization considered above
has the objective function quadratic on partially
unitary operator $u_{jk}$. There are
other objective functions
that are quadratic on partially
unitary operator $u_{jk}$ hence all the optimization above
can be applied to them as well.
With (\ref{fProjXDifferently}) definition one can consider it 
not as probability amplitude mapping $\psi_{\mathbf{x}}\to\psi_{\mathbf{f}}$, 
but as plain value mapping $\mathbf{x}\to\mathbf{f}$.
This is essentially (\ref{probgUypkAppAS}) without a denominator.
Consider
\href{https://en.wikipedia.org/wiki/Reproducing_kernel_Hilbert_space}{reproducing kernel}
$\sum_{j,j^{\prime}=0}^{m-1}f_j G^{\mathbf{f};\,-1}_{jj^{\prime}} g_{j^{\prime}}$,
it has a maximum at $\mathbf{f}=\mathbf{g}$,
assume $\mathbf{g}$ is taken from (\ref{fProjXDifferently}),
and sum it squared; obtain
\begin{align}
\mathcal{F}&=
\sum\limits_{s,s^{\prime}=0}^{m-1}\sum\limits_{k,k^{\prime}=0}^{n-1}
u_{sk}S_{sk;s^{\prime}k^{\prime}}u_{s^{\prime}k^{\prime}}
=
\sum\limits_{l=1}^{M} 
\left\lgroup\sum\limits_{j,j^{\prime}=0}^{m-1} f_j^{(l)} G^{\mathbf{f};\,-1}_{jj^{\prime}} \sum\limits_{k=0}^{n-1}u_{j^{\prime}k^{\prime}}
x_{k^{\prime}}^{(l)}\right\rgroup^2
\omega^{(l)}
\xrightarrow[u]{\quad }\max
\label{functValue} \\
S_{sk;s^{\prime}k^{\prime}}&=
\sum\limits_{j,j^{\prime}=0}^{m-1}
\Braket{f_{j}x_{k}f_{j^{\prime}}x_{k^{\prime}}}
G^{\mathbf{f};\,-1}_{js}
G^{\mathbf{f};\,-1}_{j^{\prime}s^{\prime}}
\label{SmatrixASValue}
\end{align}
This creates a different version of $S_{jk;j^{\prime}k^{\prime}}$,
a fourth order moments--type (\ref{SmatrixASValue})
instead of
previously used 
Christoffel functions product tensor $S_{sk;s^{\prime}k^{\prime}}$ from (\ref{SmatrixAS});
an important feature of (\ref{SmatrixASValue})
is that an  application of secondary sampling technique
is not required for it's calculation.

In this setup the conditions on $\Braket{f_j f_{j^{\prime}}}$ and $\Braket{x_k x_{k^{\prime}}}$
are put into the constraints (\ref{optimmatrixConstraintAppendixNUAS})
and the $\Braket{f_j x_k f_{j^{\prime}}  x_{k^{\prime}}}$
is put into the objective function\footnote{
In (\ref{functValue}) the scalar product of $f_j$ and $\sum_{k=0}^{n-1}u_{jk}x_k$
is squared and then averaged over the sample.
In finding the contributing subspace (\ref{phiiD}) it is averaged over the sample
and then squared.
This means the contributing subspace model assumes the factoring
$\Braket{f_j x_k f_{j^{\prime}}x_{k^{\prime}}}=
\Braket{f_j x_k}\Braket{f_{j^{\prime}}x_{k^{\prime}}}$.
It is similar to Lebesgue quadratures \cite{ArxivMalyshkinLebesgue},
where interchanging of averaging and taking square produces new result.
}.
The mapping with this new $S_{sk;s^{\prime}k^{\prime}}$ maps the values,
not the probabilities, but the values are considered to belong to
some vector space. The squared term in (\ref{functValue})
is just a scalar product of two vectors. With (\ref{probgUypkAppASimportantOnly})
normalizing both vectors be of unit length and the maximal
value of the objective function is $\Braket{1}$.
In (\ref{functValue}) the vectors do not have this normalizing.
One can also consider a ``partially normalized'' tensor,
the one with only $K^{(\mathbf{f})}$
term in (\ref{ChristoffelfunctionsProductMoments})
assuming ``average''--type normalizing for $x_k$ is
due to (\ref{optimmatrixConstraintAppendixNUAS}).
\begin{align}
S_{sk;s^{\prime}k^{\prime}}&=
\sum\limits_{j,j^{\prime}=0}^{m-1}
\Braket{f_{j}x_{k}|K^{(\mathbf{f})}|f_{j^{\prime}}x_{k^{\prime}}}
G^{\mathbf{f};\,-1}_{js}
G^{\mathbf{f};\,-1}_{j^{\prime}s^{\prime}}
\label{SmatrixASValueKf}
\end{align}

\subsection{\label{Optinmization}Optimization Problem}
The problem
of finding the \textsl{Knowledge Generalizing Operator}
is now reduced to maximizing (\ref{allProjUKxfAS}) coverage $\mathcal{F}$
(defined by the tensor $S_{sk;s^{\prime}k^{\prime}}$ of diverse possible forms)
subject to (\ref{optimmatrixConstraintAppendixNUAS}) constraints; the meaning of the constraints is to preserve scalar product (Gram matrix).
The result is $u_{jk}$ matrix, $j=0\dots m-1; k=0\dots n-1$.
This operator, given some input state (such as localized state $\Ket{\psi_{\mathbf{x}}}$),
uniquely (within a phase) finds the function
in $\mathbf{f}$-space
$\Ket{\mathcal{U}|\psi_{\mathbf{x}}}$
(coefficients $\alpha_j$)
that predicts
the probability (\ref{probgUypkAppAS})
of outcome $\Ket{\psi_{\mathbf{f}}}$:
\begin{align}
P(\mathbf{f})\Big|_{\mathbf{x}}&=
\Braket{\psi_{\mathbf{f}}|\mathcal{U}|\psi_{\mathbf{x}}}^2
= \frac{\left[\sum\limits_{j=0}^{m-1} \alpha_jf_j\right]^2}
{
\sum\limits_{j,j^{\prime}=0}^{m-1}f_jG^{\mathbf{f};\,-1}_{jj^{\prime}}f_{j^{\prime}}
} \label{probFXUpExpanded}
\end{align}
the $\mathbf{f}$ is
equal to the value of the outcome
we are interested to determine the probability of.
Given $\mathbf{x}$ the probability of some outcome  $\mathbf{f}$
is a squared linear function on $f_j$ multiplied by Christoffel function.

If, however, not the probability but the value of the outcome
is required --- the easiest method to obtain it is
to consider all possible $\mathbf{f}$ to
find the maximum\footnote{
\label{fcalculationFromConst}
The probability (\ref{probFXUpExpanded}) is invariant
with respect to $f_j\to const\cdot f_j$ for an arbitrary non--zero $const$.
Actual values of $f_j$ are determined using
the requirement that
the constant has always to be present
in $\mathbf{x}$- and $\mathbf{f}$- bases.
Since the value of $f_j$ corresponding to this specific index
${j:{f_j=const}}$ is always known (a constant),
the actual values of all $f_j$
are obtained as $f_j/C$ where $C=f_{j:{f_j=const}}$;
see
\texttt{\seqsplit{com/polytechnik/utils/KGOSolutionVectorXVectorF.java:evaluateAt(double [] xorig)}}.
} of (\ref{probFXUpExpanded}):
\begin{align}
\mathbf{f}&:
\max\limits_{\mathbf{f}} P(\mathbf{f})\Big|_{\mathbf{x}}=
\max\limits_{f_j} \frac{\left[\sum\limits_{j=0}^{m-1} \alpha_jf_j\right]^2}
{
\sum\limits_{j,j^{\prime}=0}^{m-1}f_jG^{\mathbf{f};\,-1}_{jj^{\prime}}f_{j^{\prime}}
}
\label{ValueF}
\end{align}
For 1D class label, where $f_j=f^j$, the problem is reduced
to finding the roots of a polynomial.
In general case the problem can be considered as generalized eigenvalue
problem with the matrices $\alpha_j\alpha_{j^{\prime}}$
(a \href{https://en.wikipedia.org/wiki/Dyadics#Dyadic,_outer,_and_tensor_products}{dyadic product} of two vectors)
and $G^{\mathbf{f};\,-1}_{jj^{\prime}}$ in the left- and right- hand sides.
It has a single non-zero eigenvalue (\ref{PmaxValue}) (equals to the maximal
probability),
corresponding eigenvector (\ref{f_Pmax})
gives the most probable outcome $\mathbf{f}$.
The maximal probability of the outcome corresponds to
the value 
\begin{align}
f^{\max P}_j&=
\sum\limits_{j^{\prime}=0}^{m-1} G^{\mathbf{f}}_{jj^{\prime}}\alpha_{j^{\prime}}
\label{f_Pmax} \\
P(\mathbf{f}^{\max P})\Big|_{\mathbf{x}}&=\sum\limits_{j,j^{\prime}=0}^{m-1}
\alpha_{j}G^{\mathbf{f}}_{jj^{\prime}}\alpha_{j^{\prime}}
\label{PmaxValue}
\end{align}
The $P(\mathbf{f}^{\max P})\Big|_{\mathbf{x}}$
is a certainty of the outcome,
the maximal possible value of (\ref{probFXUpExpanded}),
a $[0:1]$ bounded function.
A difficulty with this approach is that if $\mathbf{f}$ is constructed
from a scalar function, such as $f_j=f^j$, this relation may not hold exactly
in the result.

Obtained probability formula (\ref{probFXUpExpanded}) is of very general
form: a linear function on $f_j$ squared divided by a quadratic form on $f_j$.
It can be obtained from many different considerations,
the difference between models is in coefficients $\alpha_j$.
The simplest solution of this type
is
a ``direct projection'' solution of
\cite{malyshkin2019radonnikodym},
where we take
least squares expansion of $\Ket{f_j}$ in $\Ket{x_k}$ (\ref{fxapproxLS})
and substitute obtained $\mathbf{f}_{LS}(\mathbf{x})$
as the localization point in
(\ref{psiGlocalized})
to obtain
$\Ket{\psi_{\mathbf{f}_{LS}(\mathbf{x})}}$.
This is an example to obtain the
probability of (\ref{probFXUpExpanded}) form
without quantum channel used.

The problem
 has remarkable invariance features.
Consider (\ref{mlproblemVectorPSI}) mappings
$\psi_{\mathbf{x}^{(l)}}\to\psi_{\mathbf{f}^{(l)}}$, $l=1\dots M$ 
of $n$-dimensional vector
$\psi_{\mathbf{x}^{(l)}}$
to $m$-dimensional vector $\Ket{\psi_{\mathbf{f}^{(l)}}}$.
The vectors are projected to each other with operator $\|\mathcal{U}\|$,
projection absolute value is then squared
and all summed (\ref{allProjUKxfAS}) over the entire sample.
The major difference from any observable value--mapping technique
is that if we multiply all 
$\psi_{\mathbf{x}^{(l)}}$ and $\psi_{\mathbf{f}^{(l)}}$
by random phases $\exp(i\varphi^{(l)})$ the result will be identical!
This is the same as in quantum mechanics: a wavefunction is defined within a phase,
wavefunction absolute value squared defines the probability,
but Schr\"{o}dinger equation is written for the wavefunction.
Similarly, the knowledge generalizing operator $\|\mathcal{U}\|$
is defined (for complex matrix) within a phase,
for real matrix -- within a $\pm 1$ factor,
but the probability (\ref{probFXUpExpanded})
and coverage (\ref{allProjUKxfAS})
are equal to operator $\|\mathcal{U}\|$ projections \textsl{squared};
individual $\psi_{\mathbf{x}^{(l)}}$ and $\psi_{\mathbf{f}^{(l)}}$
may have arbitrary phases.

Optimization problem (\ref{allProjUKxfAS}) subject to (\ref{optimmatrixConstraintAppendixNUAS}) constraints is a variant of \href{https://en.wikipedia.org/wiki/Quadratically_constrained_quadratic_program}{QCQP} problem.
It has the form: to find an operator $\|\mathcal{U}\|$
optimally transforming an 
\emph{IN} state $\Ket{\psi_{\mathbf{x}}}$
into an 
\emph{OUT} state $\Ket{\psi_{\mathbf{f}}}$
on (\ref{mlproblemVectorPSI}) data,
i.e. the ideology is similar to the one of 
\href{https://en.wikipedia.org/wiki/S-matrix#Definition}{S-Matrix}.
Currently we can solve this optimization problem only numerically.
The problem  is similar to an eigenvalue problem, see (\ref{variatelagrangetovariate}). This is a new algebraic problem:
\begin{align}
  S \mathcal{U} &= \lambda \mathcal{U}
  \label{eigenvaluesLikeProblem}
\end{align}
where $S$ is a Hermitian tensor, 
``eigenvector'' $\|\mathcal{U}\|$ is a partially unitary $m\times n$ matrix,
and ``eigenvalues''  $\lambda$ is a Hermitian $m\times m$ matrix;
functional (\ref{allProjUKxfAS}) extremal value is equal to  $\lambda$
\href{https://en.wikipedia.org/wiki/Trace_(linear_algebra)}{spur}
(the sum of diagonal elements (\ref{Fextremal})). The mathematical structure of this
eigenvalue--like problem,
an ``eigenoperator'' problem,
requires a separate study and we hope
to obtain important new results soon. Currently --- we have a fast, stable to degeneracy
iteration 
algorithm to find a solution numerically,
see Appendix \ref{numAlgorithm} below.

Considered model
assumes the dynamics is determined by a single unitary operator,
possibly partially unitary.
For a $\mathbf{x}$-localized pure state
$\|\rho_{\mathbf{x}}\|=\Ket{\psi_{\mathbf{x}}}\Bra{\psi_{\mathbf{x}}}$
a  unitary operator $\|\mathcal{U}\|$ transforms the density matrix to
\begin{align} 
\|\widetilde{\rho}_{\mathbf{x}}\|&=\|\mathcal{U}|\rho_{\mathbf{x}}|\mathcal{U}^{\dagger}\| \label{Uevolution}
\end{align}
Whereas in  quantum mechanics 
evolution operator $\|\mathcal{U}\|$
corresponds to the Hamiltonian of the system:
$\mathcal{U}=\exp \left[-i\frac{t}{\hbar} H \right]$,
in data analysis knowledge generalizing operator $\|\mathcal{U}\|$ is obtained
from optimization problem (\ref{allProjUKxfAS})
subject to (\ref{optimmatrixConstraintAppendixNUAS}) constraint.
Quantum evolution of (\ref{Uevolution}) form always transforms
a pure state $\|\rho\|=\Ket{\psi}\Bra{\psi}$
to the pure state $\|\widetilde{\rho}\|=\Ket{\mathcal{U}|\psi}\Bra{\psi|\mathcal{U}^{\dagger}}$,
and a mixed state $\|\rho\|$ 
to the mixed state $\|\widetilde{\rho}\|$.
In data analysis there is a common situation
when a pure state is transformed into a mixed state,
Markov chain is an example. In this case a more general form of
quantum evolution is required\cite{kraus1983states}:
\begin{align}
  \widetilde{\rho}&=\sum\limits_s B_s\rho B^{\dagger}_s \label{KrausOperator}
\end{align}
with
\href{https://en.wikipedia.org/wiki/Quantum_operation#Kraus_operators}{Kraus operators}
$B_s$ satisfying\footnote{
Similarly to (\ref{optimmatrixConstraintAppendixNUAS}) Kraus operators $B_s$
can  also be considered in a ``partially unitary''--style
with $b_{s;jk}$
matrices of the dimension $m\times n$ satisfying
$\Braket{f_j|f_{j^{\prime}}}= \sum\limits_{s}\sum\limits_{k,k^{\prime} =0}^{n-1}b_{s;jk}
\Braket{x_k|x_{k^{\prime}}}
b^*_{s;j^{\prime}k^{\prime}}$.
The optimization problem (\ref{allProjUKxfAS})
then becomes
$\sum\limits_s\sum\limits_{l=1}^{M}
\Braket{\psi_{\mathbf{f}^{(l)}}|B_{s}|\psi_{\mathbf{x}^{(l)}}}^2
\omega^{(l)}
\xrightarrow[B_{s}]{\quad }\max$.
}
\begin{align}
  \sum\limits_s B_s^{\dagger}B_s&=\mathds{1} \label{constraintKrauss}
\end{align}
The data we use in this
paper is of pure state to pure state mapping (\ref{mlproblemVectorPSI}).
For other type of input data unitary evolution (\ref{Uevolution}) should be
replaced by a more general form (\ref{KrausOperator});
one may think about it as a quantum system evolving with several
Hamiltonians at once
$B_s=\exp \left[-i\frac{t}{\hbar} H_s \right]$,
not as about a system evolving with the Hamiltonian $H=\sum\limits_s H_s$.
The approach is directly generalizable to e.g. probability
distribution to probability distribution mapping:
in this case the observations are not
localized states mapping $\psi_{\mathbf{x}^{(l)}}\to\psi_{\mathbf{f}^{(l)}}$,
but corresponding density matrices mapping
$\|\rho_{\mathbf{x}}^{(l)}\|\to\|\rho_{\mathbf{f}}^{(l)}\|$.

Initial $\mathbf{x}^{(l)}\to\mathbf{f}^{(l)}$ input data (\ref{mlproblemVector})
was converted to pure state to pure state mapping
$\psi_{\mathbf{x}^{(l)}}\to\psi_{\mathbf{f}^{(l)}}$
 (\ref{mlproblemVectorPSI})
to formulate optimization problem (\ref{allProjUKxfAS}) 
subject to (\ref{optimmatrixConstraintAppendixNUAS}) constraints.
It is essential from methodical point of view
to discuss what input moments are required for this problem (to obtain
the tensor $S_{jk;j^{\prime}k^{\prime}}$ (\ref{SmatrixAS})) 
and 
compare with other models.
This is summarized in the table:
\begin{center}
\begin{tabular}{||>{\raggedright}p{6.3cm}|p{8.5cm}||}
\hhline{|t:=:=:t|}
Model & Tensors Required to Calculate \\[0.5ex]
\hhline{||-|-||}
Least Squares (\ref{fxapproxLS}) & $\Braket{x_kx_{k^{\prime}}}$, $\Braket{x_kf_j}$ \\
Radon-Nikodym (\ref{RNfsolutionX}) & $\Braket{x_kx_{k^{\prime}}}$, $\Braket{x_kx_{k^{\prime}}f_j}$ \\
$\mathbf{x}$ --- $\mathbf{f}$
Christoffel function (\ref{zDef}) &
 $\Braket{x_kx_{k^{\prime}}}$,
 $\Braket{f_jf_{j^{\prime}}}$, $\Braket{x_kf_j}$ \\
Pure Joint Distribution (\ref{probgypkApp}) &
 $\Braket{x_kx_{k^{\prime}}}$,
 $\Braket{f_jf_{j^{\prime}}}$, $\Braket{x_kf_j}$,
$\Braket{x_{k}f_{j}|K^{(\mathbf{x})}K^{(\mathbf{f})}|x_{k^{\prime}}f_{j^{\prime}}}$
\\
 Partial Unitarity (KGO) (\ref{probgUypkAppAS}) &
  $\Braket{x_kx_{k^{\prime}}}$, $\Braket{f_jf_{j^{\prime}}}$,
$\Braket{x_{k}f_{j}|K^{(\mathbf{x})}K^{(\mathbf{f})}|x_{k^{\prime}}f_{j^{\prime}}}$
\\
Partial Unitarity (KGO) $K^{(\mathbf{f})}$
(\ref{SmatrixASValueKf})  &
  $\Braket{x_kx_{k^{\prime}}}$, $\Braket{f_jf_{j^{\prime}}}$,
$\Braket{x_{k}f_{j}|K^{(\mathbf{f})}|x_{k^{\prime}}f_{j^{\prime}}}$
\\
 Partial Unitarity (KGO) adj. (\ref{probgUypkAppASimportantOnly})
  &
  Beyond moments, no $\Braket{x_kf_j}$ used.
\\
\hhline{|b:=:=:b|}
\end{tabular}
\end{center}
The major difference --- Knowledge Generalizing Operator (KGO)
is the only model that
does not require ``projective'' moments
$\Braket{x_kf_j}$;
it requires
Gram matrices (\ref{GramX}) and (\ref{GramF})
of \emph{IN} and \emph{OUT} bases
and
the moments of
the Christoffel functions product (\ref{ChristoffelfunctionsProductMoments}). 
These moments can be obtained with
an  application of secondary sampling technique\cite{MalMuseScalp}:
Gram matrices are built first;
then,
for every observation $l=1\dots M$,
Christoffel function is calculated
and used
as it were plain observed at observation $l$.
These moments\footnote{
The  (\ref{probgUypkAppASimportantOnly}) KGO model goes
``beyond moments''. Even with secondary sampling it is impossible
to build from moments the (\ref{allProjU}) target functional 
with the probability (\ref{probgUypkAppASimportantOnly}).
Moreover, this problem is not a QCQP problem.
} of two Christoffel functions product 
are the input used to formulate the problem (\ref{allProjUKxfAS}).
For a Christoffel function 
in some multi-dimensional vector space $\bm{r}$
(e.g. $\mathbf{x}$ (\ref{ChristoffelFunctionDef})
  or $\mathbf{f}$ (\ref{ChristoffelF}))  
with $\Braket{\cdot}$ inner product and non-degenerated Gram matrix
$G_{jj^{\prime}}=\Braket{r_j|r_{j^{\prime}}}$
there is a $1/r^2$ asymptotic:
\begin{align}
K(\bm{r})&=
\frac{1}{
\Braket{\bm{r}|G^{-1}|\bm{r}}
}=\frac{1}{
\sum\limits_{j,j^{\prime}}r_jG^{-1}_{jj^{\prime}}r_{j^{\prime}}
}
\label{ChristoffelFunctionDefGeneral} \\
K(\bm{r})&\sim 1/r^2 &
\text{for }\enspace r\to\infty
\label{CoulombLaw}
\end{align}
The same $1/r^2$ long--range interaction
presents
in
\href{https://en.wikipedia.org/wiki/Coulomb%27s_law}{Coulomb's law}
or
\href{https://en.wikipedia.org/wiki/Newton%27s_law_of_universal_gravitation}{Newton's law of gravitation}.
With (\ref{CoulombLaw}) asymptotic
the Christoffel function can be viewed as a form of ``long--range $1/r^2$ interaction'', an anisotropic gravity--like law of data analysis.
These non-local features, along with eigenproblem
(\ref{lagrangetovariateNUDlen}) 
of the dimension $Dn$
and  SVD (\ref{SVDNUDlen}) (or Gram matrix eigenproblem (\ref{GramEVtransform}))
that are required on every iteration,
substantially slow down the algorithm when implemented without optimization.
At this point, however, the goal is not to build a fast algorithm,
but to understand all the benefits and drawbacks
of ML knowledge representation in the form
  of partially unitary operator. Let us do a demonstration.

\section{\label{ADemonstration}A Demonstration Of Knowledge Generalizing Operator Application}
In this section we are going to present several demonstrations
of $\mathbf{f}(\mathbf{x})$ calculation using (\ref{ValueF}).
The $\mathbf{f}$ and $\mathbf{x}$ are treated as linear spaces,
a basis for wavefunction, with partially unitary operator $u_{jk}$
mapping (\ref{fProjXDifferently}).
The result is invariant relatively $\mathbf{f}\to C \cdot \mathbf{f}$.
To obtain actual value of $\mathbf{f}$ --- it should be
\hyperref[fcalculationFromConst]{normalized to const}.
The constant has always to be present in both  $\mathbf{f}$-- and $\mathbf{x}$--
bases. Thus
\begin{align}
\mathbf{f}^{actual}&=\frac{\mathbf{f}}{f_{j:{f_j=const}}}
\label{fjconst}
\end{align}
In this equation the nominator is a linear function on $\mathbf{x}$
(\ref{fProjXDifferently})
and the denominator, the const--component of $\mathbf{f}$,
possibly also is a linear function on $\mathbf{x}$.
Thus the value obtained from partially unitary
operator mapping is a ratio of two linear functions on $\mathbf{x}$.
The least squares (\ref{fxapproxLS}) always maps a constant  to a constant,
thus when $u_{jk}$ is a least squares mapping the denominator in (\ref{fjconst})
is always a constant. In Radon--Nikodym mapping (\ref{RNfsolutionX})
the nominator is a quadratic form on $\mathbf{x}$ and the denominator
is a positive quadratic form on $\mathbf{x}$; the denominator is never zero.
In (\ref{fjconst}) the nominator and the denominator
are both linear functions on $\mathbf{x}$ of most general form.
The divergences coming from denominator's zeroes
are important new features of the approach. In least squares --
these zeroes are on the infinity.  Denominator's zeroes may come either
from deep internal properties of the model 
or
from sub-optimal
solution of the optimization problem (or badly chosen objective function).

The objective function is determined by the
tensor $S_{jk;j^{\prime}k^{\prime}}$. Whereas properly normalized
probability (\ref{probgUypkAppASimportantOnly}) lead to
a non--QCQP problem,
the original Christoffel (\ref{SmatrixAS}),
the adjusted number of degrees of freedom Christoffel (\ref{ChristoffelfunctionsProductMomentsAdjusted}),
$\mathbf{f}$--Christoffel (\ref{SmatrixASValueKf}),
and plain $\Braket{f_{j}x_{k}f_{j^{\prime}}x_{k^{\prime}}}$ (\ref{SmatrixASValue})
have $S_{jk;j^{\prime}k^{\prime}}$ tensor readily available
and the optimization problem (\ref{allProjUKxfAppendix})
with the constraints (\ref{optimmatrixConstraintAppendixNUAppendix})
can be formulated and solved numerically.

Among available $S_{jk;j^{\prime}k^{\prime}}$ versions the
$\mathbf{f}$--Christoffel (\ref{SmatrixASValueKf})
has the most ``usual'' properties. For example
the (\ref{SmatrixAS}) or (\ref{ChristoffelfunctionsProductMomentsAdjusted}),
when run with a data of exact $\mathbf{x}\to\mathbf{f}$
\href{https://en.wikipedia.org/wiki/Homomorphism}{homomorphism}
can possibly give a higher $\mathcal{F}$ on non--exact mapping due to
unusual localized states normalizing.
For this reason all the demonstrations below will be performed
with $\mathbf{f}$--Christoffel $S_{jk;j^{\prime}k^{\prime}}$ (\ref{SmatrixASValueKf}).

\begin{figure}[t]
\includegraphics[width=16cm]{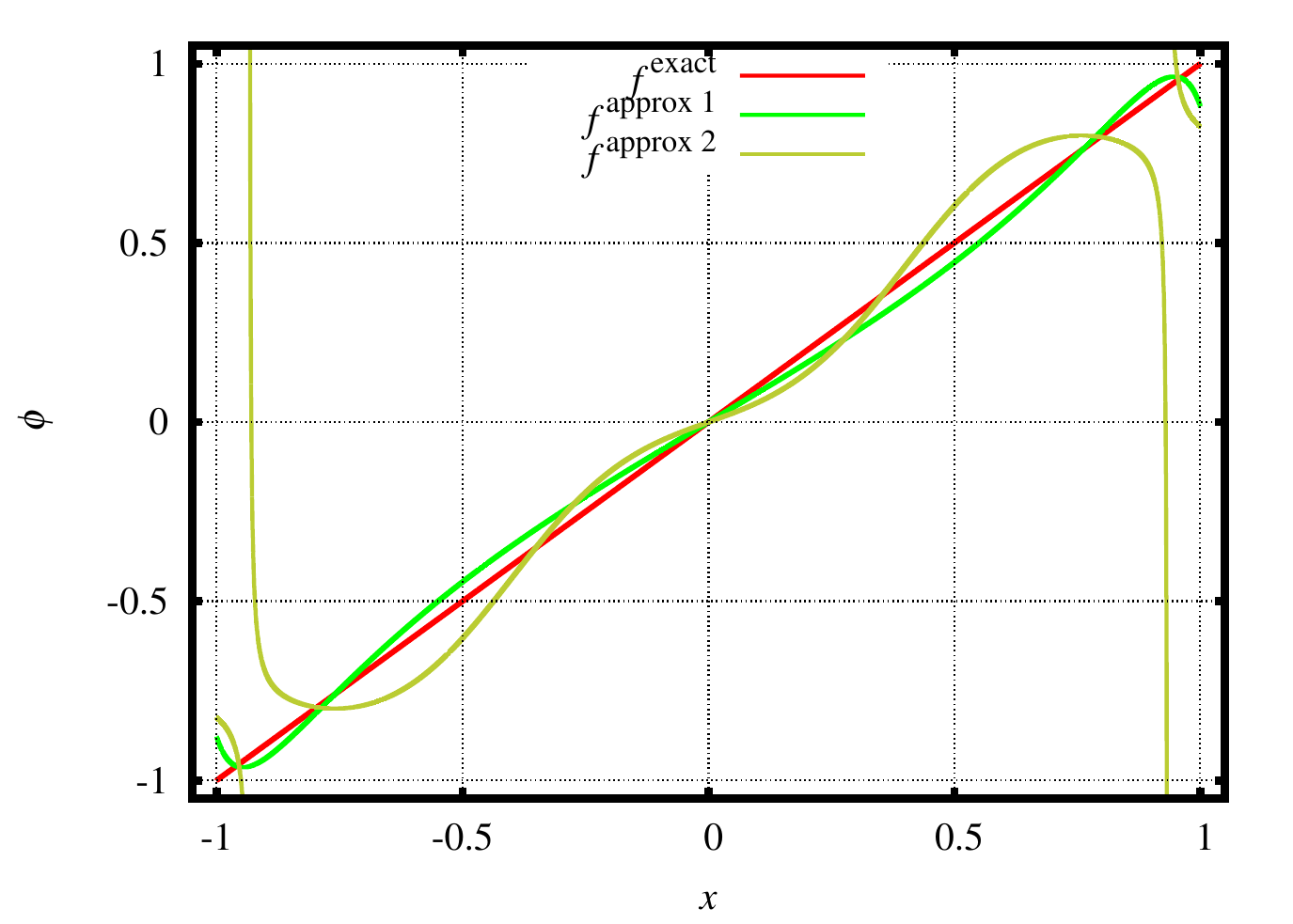}
\caption{\label{FnonExact}
For a data with known exact $f=x$ solution,
when numerical method does not find it -- it is possible to have
zeroes in (\ref{fjconst}) and corresponding poles in the behavior.
}
\end{figure}
Consider a trivial mapping with 
the measure $\Braket{g}=\int_{-1}^{1}g(x)dx$
and the basis $\mathbf{x}$ constructed from 1D variable $x\in [-1:1]$
as  $x_k=x^k$, $k=0\dots 6; n=7$, and $f_j=x_j$ for $j=0\dots 4; m=5$.
The solution is trivial: take first $m$ components of $x_k$ and regularize;
then use them for both: $\mathbf{x}$ and $\mathbf{f}$.
However, when the numerical algorithm cannot find this exact solution
we can observe a deviation from exact match.
In Fig. \ref{FnonExact}
the exact solution  along with two approximate solutions
of different quality are presented.
A not very accurate  approximate numerical solution may give poles corresponding to
the
zeroes in (\ref{fjconst}) denominator (clearly observed for $f^{\mathrm{approx}\, 2}$ near interval edge).

\begin{figure}
\includegraphics[width=16cm]{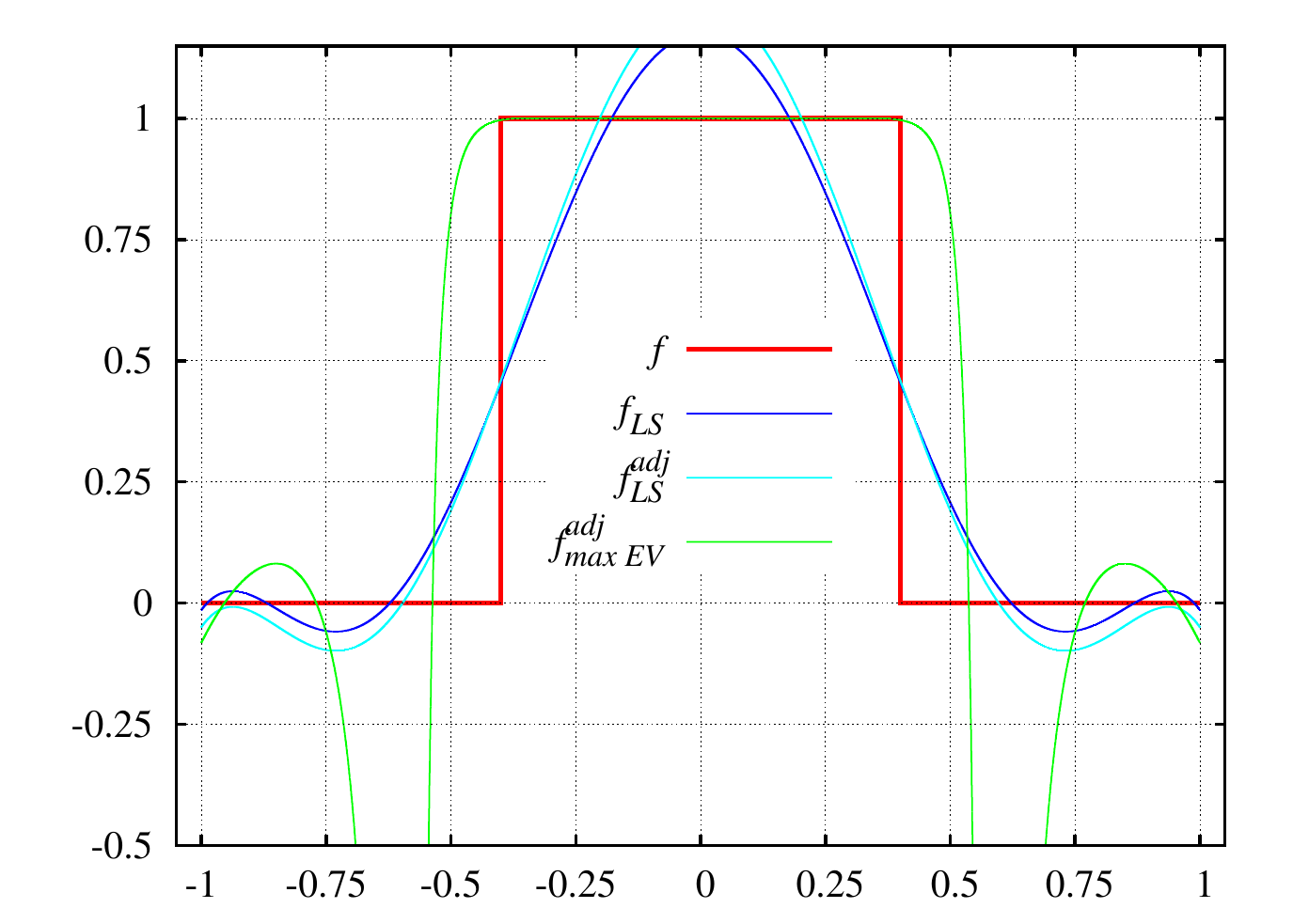}
\caption{\label{UstepU}A square wave step function (the same as in Fig. \ref{RNapproximationSquareWave}), with least squares (blue), least squares with (\ref{SVDadjustmentSimple}) SVD adjustment (light blue), and maximal eigenvalue with (\ref{SVDadjustmentSimple}) SVD adjustment (green).
}
\end{figure}

In Fig. \ref{UstepU} a square wave step function (the same as in Fig. \ref{RNapproximationSquareWave}) is presented with the same measure and basis; $n=7$.
The $f$ takes only two values since the only available $m$ is $m=2$.
The exact solution was difficult to obtain numerically as the problem is
substantially degenerated. We present three approximate solutions.
The blue line
is regular least squares (\ref{fxapproxLS}). Light blue is the same least squares mapping
(\ref{fxapproxLS})
adjusted with (\ref{SVDadjustmentSimple}) to partial unitarity. Green --- maximal eigenvalue
(\ref{allProjUKxfAppendixDIAG})
solution adjusted to partial unitarity with (\ref{SVDadjustmentSimple}).
One can see that partial unitarity adjustment makes little changes
to least squares solution. For adjusted maximal eigenvalue solution
the (\ref{fjconst}) denominator poles are close to the support
of $\mathbf{x}$, this creates two artifacts in $\mathbf{f}$.
Note almost exact $f=1$ matching in the center.

Consider a 2D example. Let us take an image
and consider it as a two--dimensional basis
mapping a pixel coordinate $(x,y)$ to gray intensity $f$.
\begin{align}
  (x,y)^{(l)}&\to f^{(l)} & & \text{weight $\omega^{(l)}=1$}  \label{lenmlproblem} \\
  \mathbf{k}&=(k_x,k_y) 
\label{lenMI} \\
x_{\mathbf{k}}&=x^{k_x}y^{k_y}
& &0 \le k_x \le n_x-1 ;{\quad }0 \le k_y \le n_y-1
\label{Ximage} \\
  f_j&=f^j & & j=0\dots m-1 \label{Fimage}
\end{align}
This forms a (\ref{mlproblemVector}) basis\footnote{
For numerical stability it is better to use argument--scaled Chebyshev polynomials
rather than monomials powers $x^{k_x}y^{k_y}$ and $f^j$.}
of $n=n_x n_y$ and $m$
dimensions. Let us construct
an operator $u_{jk}$ mapping $\mathbf{x}\to\mathbf{f}$.
A simple example is least squares (\ref{fxapproxLS}),
it creates a familiar image expansion similar to Fourier series.
However, we are interested in operators $u_{jk}$ satisfying
all partial unitarity constraints (\ref{optimmatrixConstraintAppendixNUAS}).
A simple  variant of constraint--satisfying operator
can be obtained from any $u_{jk}$ operator applying Appendix \ref{numAlgorithmApproximate}
algorithm.
\begin{figure}[t]
\includegraphics[width=5.2cm]{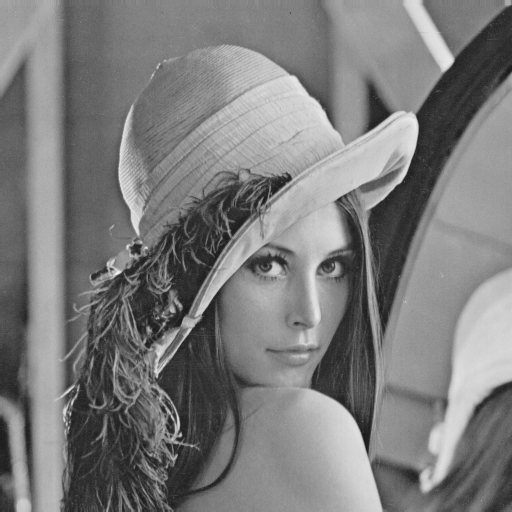}
 \includegraphics[width=5.2cm]{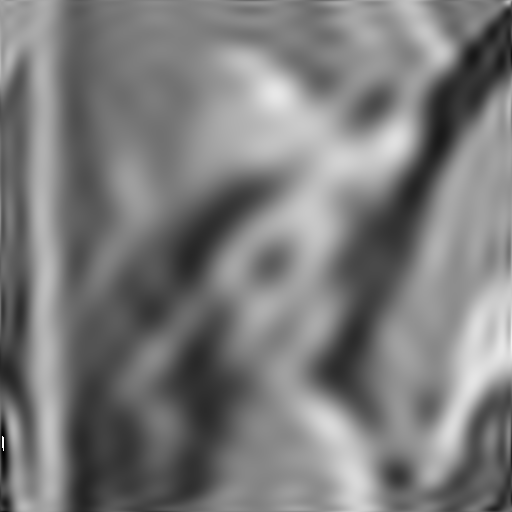}
 \includegraphics[width=5.2cm]{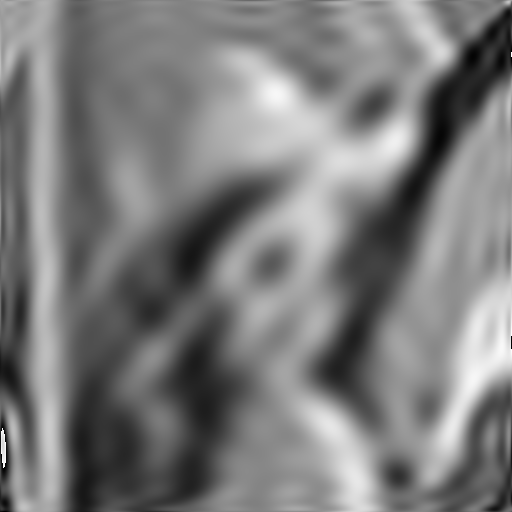} \\
 \includegraphics[width=5.2cm]{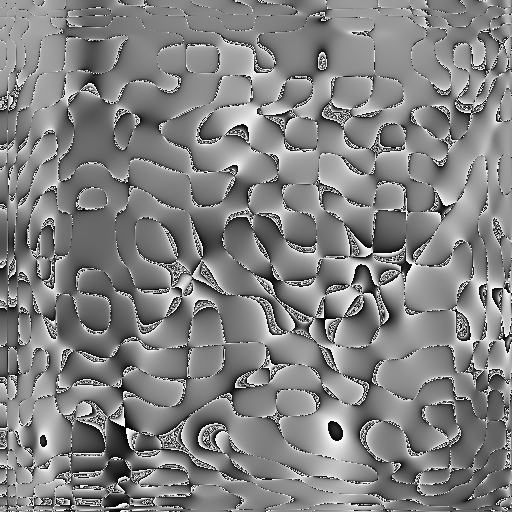}
 \includegraphics[width=5.2cm]{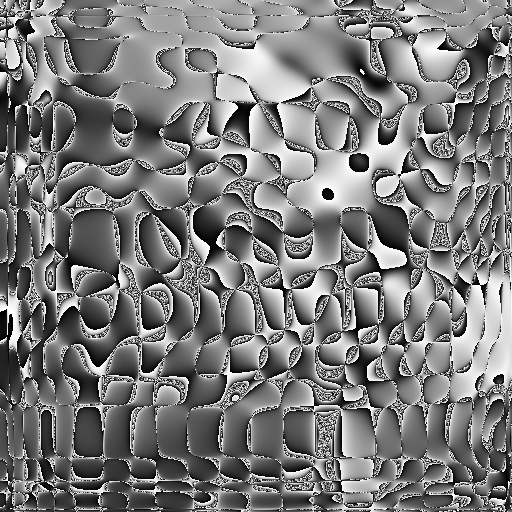}
 \includegraphics[width=5.2cm]{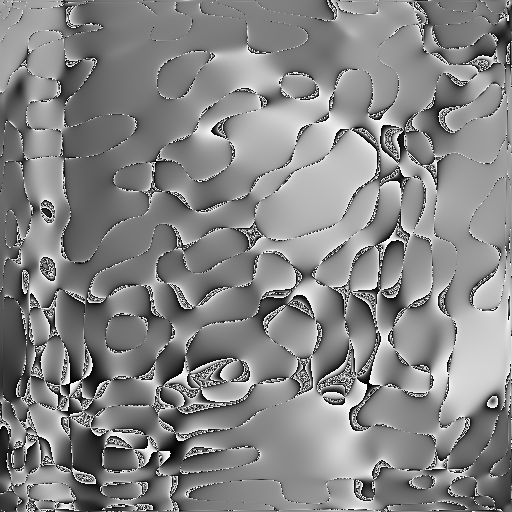} \\
\includegraphics[width=5.2cm]{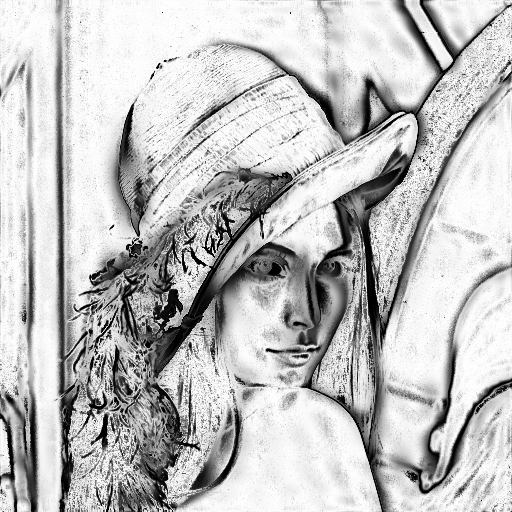}
\includegraphics[width=5.2cm]{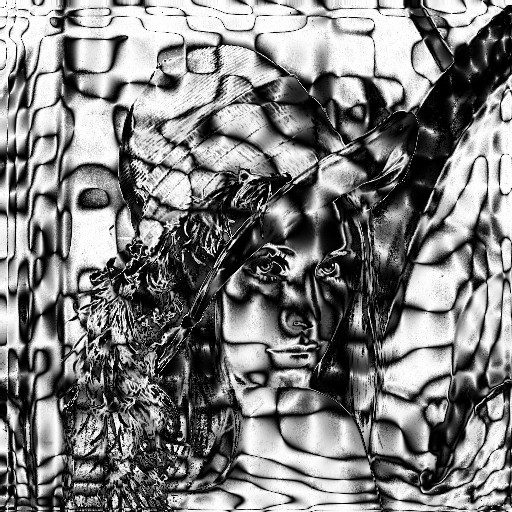}
\includegraphics[width=5.2cm]{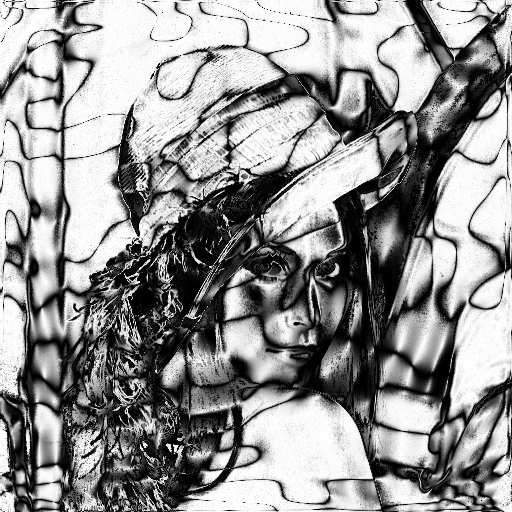}
 \caption{\label{lenna}
 \linespread{1.0}\selectfont{}
A demonstration of image interpolation with
 $n_x=n_y=25$, $m=5$.
 Top row: original image, least squares (\ref{fxapproxLS}) interpolated,
 and the same least squares adjusted with (\ref{SVDadjustmentSimple})
 to partial unitarity constraints.
 Middle row: optimization (\ref{allProjUKxfAS}) with simplified constraints (\ref{optimmatrixConstraintScalarAppendix}) (the state of maximal eigenvalue),
the same one
adjusted with (\ref{SVDadjustmentSimple})
 to partial unitarity constraints,
and optimization result with Section \ref{linearConstraintOpt} algorithm.
Bottom row: The probability (\ref{probFXUpExpanded})
is calculated at actual $\mathbf{f}$, white $P=1$, black $P=0$.
It is calculated
for: least squares (\ref{fxapproxLS})
(``direct projection'' model of \cite{malyshkin2019radonnikodym}),
the state of maximal eigenvalue (unadjusted),
and Section \ref{linearConstraintOpt} algorithm.
}
\end{figure}
In Fig. \ref{lenna} (top row) we present original image, least squares expansion
and constraint--adjusted least squares for $n_x=n_y=25$, $m=5$.
The constraint--adjusted least squares is very similar to the original least squares.
The least squares operator maps pixel coordinates to gray intensity,
not the localized states wavefunction. When an operator is optimized
to map the wavefunctions this may cause poles in values, the zeroes
of (\ref{fjconst}) denominator. It is trivially to construct
a partially unitary operator $u_{jk}$ preserving the constant:
construct a partially unitary operator
mapping  $\mathbf{x}$-space \textsl{without const}
to  $\mathbf{f}$-space \textsl{without const}
$\mathbf{x} \backslash C \to \mathbf{f} \backslash C$,
then do a direct sum with $C \to C$ mapping. We do not perform
such a transform specifically to observe the poles in (\ref{fjconst}).
We present three pictures, corresponding
to $u_{jk}$ operators differently optimizing (\ref{allProjUKxfAS})
with
$\mathbf{f}$--Christoffel tensor (\ref{SmatrixASValueKf}).
In Fig. \ref{lenna} (middle row)
we present the results corresponding to these three $u_{jk}$:
optimizing (\ref{allProjUKxfAS}) with simplified constraints (\ref{optimmatrixConstraintScalarAppendix}),
the same one adjusted with
(\ref{SVDadjustmentSimple})
 to partial unitarity,
and optimization result with Section \ref{linearConstraintOpt} algorithm
(overall the best optimization algorithm we have so far).
Left two pictures in the middle row ---
a simple solutions (based on trivial approach of maximal eigenvalue state),
they have
noticeable $1/n_{\{x,y\}}$ scale artifacts.
The last one is very close to the global maximum of (\ref{allProjUKxfAS})
and  ``mixes'' the modes much stronger .
The poles of (\ref{fjconst}) separate the regions
and the structure of these ``separators'' can be a subject of our future research.

The developed approach works with probabilities, not with
the values. For this reason it is of interest
to present the probability (\ref{probFXUpExpanded})
at given known outcome $\mathbf{f}=\mathbf{f}^{(l)}$.
The result is presented in Fig. \ref{lenna},
the bottom row. The probability is scaled as white $P=1$, black $P=0$.
It is presented in the bottom row for three algorithms:
least squares (\ref{fxapproxLS}),
the state of maximal eigenvalue (unadjusted),
and Section \ref{linearConstraintOpt} algorithm.

The method to overcome 
noticeable $1/n_{\{x,y\}}$ artifacts in
Fig. \ref{lenna} is to use properly
normalized states (\ref{probgUypkAppASimportantOnly}).
In most general form it can be considered as an \textsl{unconstrained}
optimization problem.
Given sampled data (\ref{mlproblemVector})
find a linear transform $u_{jk}$ (\ref{fProjXDifferentlyGENERAL}),
a general form matrix of the dimension  $m\times n$,
maximizing (\ref{properlyNormalizedPsi})
\begin{align}
f_j&=\sum\limits_{k=0}^{n-1}u_{jk}x_k
& j=0\dots m-1
\label{fProjXDifferentlyGENERAL} \\
\mathcal{F}&=
\sum\limits_{l=1}^{M}
\Braket{\psi_{\mathbf{f}^{(l)}}|\psi_{u(\mathbf{x}^{(l)})}}^2
\omega^{(l)}
\xrightarrow[u]{\quad }\max
\label{properlyNormalizedPsi}
\end{align}
Here the $\Ket{\psi_{\mathbf{g}}}$ is
the state (\ref{psiGlocalized})
localized at $\mathbf{f}=\mathbf{g}$,
and $\Ket{\psi_{u(\mathbf{x})}}$
is also
 $\mathbf{f}$--localized state (\ref{psiGlocalized})
with
the localization point $\mathbf{g}$
determined by (\ref{fProjXDifferentlyGENERAL})
linear mapping. When expanded $\Braket{\psi_{\mathbf{g}}|\psi_{u(\mathbf{y})}}^2$
is (\ref{probgUypkAppASimportantOnly}).
The objective function (\ref{properlyNormalizedPsi})
is the total probability transferred from $\mathbf{x}$--space
to $\mathbf{f}$--space; this is an unconstrained problem.
In this most general form the problem is not
a QCQP problem and it is difficult to solve numerically;
the difficulty is that
with $\Ket{\psi_{u(\mathbf{x})}}$ state the operator $u_{jk}$
enters (through localization point $\mathbf{g}$) both the nominator
and the denominator of (\ref{psiGlocalized}),
what makes the optimization problem (\ref{properlyNormalizedPsi})
not a QCQP problem.
The problem can be substantially simplified
when the $u_{jk}$ mapping is considered to be a partially
unitary transform (\ref{optimmatrixConstraintAppendixNUAS})
to obtain a QCQP problem.
The problem can be further approximated
by splitting the solution into two steps:
selecting the contributing subspace $\phi_k$ of the dimension $m$,
then constructing a unitary (not partially unitary)
mapping from the contributing subspace to $f_j$.
A simple projective approach is presented above
in Eq. (\ref{FtotExpansion}) or, more generally, in the Appendix \ref{SubSpaceSelection} below.
A simple solution of this type is
the ``direct projection'' model of
\cite{malyshkin2019radonnikodym}
where the localization point
is determined from plain least squares (\ref{fxapproxLS})
to obtain the state $\Ket{\psi_{\mathbf{f}_{LS}(\mathbf{x})}}$.
The probability
$\Braket{\psi_{\mathbf{f}^{(l)}}|\psi_{\mathbf{f}_{LS}(\mathbf{x}^{(l)})}}^2$
of the ``direct projection'' model is presented in Fig. \ref{lenna}
(leftmost in the bottom row).

These demonstrations make us to conclude
that partial unitary mapping is a rich form of knowledge representation
with a high generalizing power, however a more study is required.

\section{\label{conclusion}Conclusion}

The developed knowledge generalizing operator concept
is similar to the
\href{https://en.wikipedia.org/wiki/S-matrix#Definition}{S-Matrix}
approach since it is an operator
optimally transforming an 
\emph{IN} state $\Ket{\psi_{\mathbf{x}}}$
into an 
\emph{OUT} state $\Ket{\psi_{\mathbf{f}}}$.
As any  wavefunction in ML
is known within an arbitrary phase
the equation for the operator must include only observable values.
The problem we consider is to
recover $\|\mathcal{U}\|$
from all it's 
 projections \textsl{squared},
from the probabilities (\ref{probgUypkAppAS}).
The condition of operator's optimality is (\ref{allProjUKxfAS}) coverage  maximization
on (\ref{mlproblemVectorPSI}) data;
it is
a new kind of algebraic problem
 (\ref{eigenvaluesLikeProblem}) ---
the equation to determine the $\|\mathcal{U}\|$.
The situation is the same as with the Schr\"{o}dinger equation:
the equation is written for $\psi$, but only $\psi^2$ is observable.
This is the difference between our and all other ML knowledge
representation techniques where knowledge representation characteristics
are observable values.
If a model relates an initial observable and the final observable
then it is a ``joint distribution model'', it cannot predict
something that has not been already observed in the training data.
Knowledge generalizing operator relates
the amplitude of the initial state to
the amplitude of the final state.
This is the very feature that creates generalization.
The same is in quantum mechanics: $\psi^2$ vs $\psi$;
whereas a mapping of $\psi^2$ is meaningless,
the mapping of $\psi$ determines the dynamics of a system.

Considered maximization problem (\ref{allProjUKxfAS})
is a simple
example of knowledge generalizing operator technique:
for observations $l=1\dots M$  convert 
$\mathbf{x}^{(l)} \to \mathbf{f}^{(l)}$
to $\psi_{\mathbf{x}^{(l)}}\to\psi_{\mathbf{f}^{(l)}}$,
then reconstruct $\|\mathcal{U}\|$ from
it's projections squared
$\Braket{\psi_{\mathbf{f}^{(l)}}|\mathcal{U}|\psi_{\mathbf{x}^{(l)}}}^2$.
The problem can be
 generalized by considering, instead of $l$, $\mathbf{x}$,
$\mathbf{f}$, and $\Braket{\cdot}$,
the structures generalizing the concepts of set, vector, and measure.
In the most general form it can be formulated as:
for $\psi\in S_x$ and $\varphi\in S_f$ recover
partially unitary operator $\|\mathcal{U}\|$
from it's projections squared
 $\sum\limits_{l\in M}\omega^{(l)}\Braket{\varphi^{(l)}|\mathcal{U}|\psi^{(l)}}^2\xrightarrow[{\mathcal{U}}]{\quad }\max$.
 The problem can be further generalized by considering
 mixed states $\|\rho\|\in S_x$ and $\|\varrho\|\in S_f$
 and recovering Kraus operators $B_s$ (\ref{KrausOperator}) from
projections squared:
$\sum\limits_{l\in M}\omega^{(l)}\sum\limits_s\mathrm{Spur}\|\varrho^{(l)}|B_s|\rho^{(l)}|B_s^{\dagger}\|
\xrightarrow[{B_s}]{\quad }\max$.

There is another interesting twist
to the considered problem of finding a partially unitary matrix
 $u_{jk}$
of the dimension
$\dim(\emph{OUT}) \times \dim(\emph{IN})$
mapping operators from $\Ket{\emph{IN}}$ to $\Ket{\emph{OUT}}$.
Consider the problem:
for  $\dim(\emph{OUT}) < \dim(\emph{IN})$
select $\dim(\emph{OUT})$ input attributes
out of all $\dim(\emph{IN})$ available
that maximize some correctness condition
which is a function of all selected attributes.
For all interesting correctness conditions
this problem
is typically a one of 
\href{https://en.wikipedia.org/wiki/NP-completeness}{NP--complete} type.
There is a single correctness function (least squares)
that can be trivially solved.
Maximization of total matched probability  (\ref{allProjU})
among all partially unitary operators $u_{jk}$
also selects $\dim(\emph{OUT})$ inputs
from all $\dim(\emph{IN})$ available.
This is a new algebraic problem (\ref{eigenvaluesLikeProblem}).
Found mapping $u_{jk}$
can be viewed as a solution to attributes selection problem
with correctness conditions
somewhere ``in between''
least squares and NP--complete,
for example there is a simple subspace selection
approach (\ref{GEVKftoXfxfXX}) --- then a problem of unitary mapping (not partially unitary)
can be directly solved.

\appendix
\section{\label{numAlgorithm}A Numerical Solution to Find
the Knowledge Generalizing Operator}
The problem we consider is a
\href{https://en.wikipedia.org/wiki/Quadratically_constrained_quadratic_program}{QCQP}
problem
to maximize (\ref{allProjUKxfAppendix})
subject to (\ref{optimmatrixConstraintAppendixNUAppendix}) constraint.
\begin{align}
\mathcal{F}&
=\sum\limits_{l=1}^{M}\Braket{\psi_{\mathbf{f}^{(l)}}|\mathcal{U}|\psi_{\mathbf{x}^{(l)}}}^2
\omega^{(l)}
=\sum\limits_{j,j^{\prime}=0}^{D-1}\sum\limits_{k,k^{\prime}=0}^{n-1}
u_{jk}S_{jk;j^{\prime}k^{\prime}}u_{j^{\prime}k^{\prime}}
\xrightarrow[{u}]{\quad }\max
\label{allProjUKxfAppendix} \\
\Braket{f_j|f_{j^{\prime}}}&= \sum\limits_{k,k^{\prime} =0}^{n-1}u_{jk}\Braket{x_k|x_{k^{\prime}}} u_{j^{\prime}k^{\prime}}& j,j^{\prime}=0\dots D-1
    \label{optimmatrixConstraintAppendixNUAppendix}
\end{align}
Without loss of generality we put
$\delta_{kk^{\prime}}=\Braket{x_k|x_{k^{\prime}}}$ and
$\delta_{jj^{\prime}}=\Braket{f_j|f_{j^{\prime}}}$
as we can always choose an orthogonal basis
by applying, for example,
an orthogonalization of
\href{https://en.wikipedia.org/wiki/Gram%E2%80%93Schmidt_process}{Gram--Schmidt}
type.
Contrary to other methods (e.g. regular
\href{https://en.wikipedia.org/wiki/Principal_component_analysis#Further_considerations}{principal components})
the result obtained with knowledge generalizing operator
is invariant with respect to (\ref{gaugeXF}) transform of input data,
thus it does not depend on initial regularization.
The problem becomes:
\begin{align}
\mathcal{F}&=
\sum\limits_{j,j^{\prime}=0}^{D-1}\sum\limits_{k,k^{\prime}=0}^{n-1}
u_{jk}S_{jk;j^{\prime}k^{\prime}}u_{j^{\prime}k^{\prime}}
\xrightarrow[{u}]{\quad }\max
\label{allProjUKxfAppendixDIAG} \\
\delta_{jj^{\prime}}&= \sum\limits_{k=0}^{n-1}u_{jk}u_{j^{\prime}k}& j,j^{\prime}=0\dots D-1
    \label{optimmatrixConstraintAppendixNUAppendixDIAG}
\end{align}
Consider
the squared
\href{https://en.wikipedia.org/wiki/Matrix_norm#Frobenius_norm}{Frobenius norm}
of matrix $u_{jk}$
to be a ``simplified constraint'':
\begin{align}
    &\sum\limits_{j=0}^{D-1}\sum\limits_{k=0}^{n-1}u^2_{jk} =D
    \label{optimmatrixConstraintScalarAppendix}
\end{align}
This is a ``partial'' constraint (it is the sum of all (\ref{optimmatrixConstraintAppendixNUAppendixDIAG}) diagonal elements).
For this ``partial'' constraint optimization problem (\ref{allProjUKxfAppendixDIAG})
can be readily converted to an eigenvalue problem that can be directly solved.
The main idea is to adjust this ``preliminary'' solution
to satisfy the full set of (\ref{optimmatrixConstraintAppendixNUAppendixDIAG})
constraints and then
calculate new values of Lagrange multipliers.
Performing several iterations the process possibly converge
to (\ref{allProjUKxfAppendixDIAG}) maximum
with all the required
constraints (\ref{optimmatrixConstraintAppendixNUAppendixDIAG})
satisfied.
In \cite{malyshkin2019radonnikodym} a similar technique has been tried
for a unitary operator (\ref{constRaintUnitarity}). The (\ref{optimmatrixConstraintAppendixNUAppendixDIAG})
corresponds to partially orthogonal operator (partially unitary real matrix):
$D\le n$.

Consider Lagrange multipliers $\lambda_{jj^{\prime}}$, a matrix
of $D\times D$ dimension,
to approach optimization problem (\ref{allProjUKxfAppendixDIAG})
with the constraints (\ref{optimmatrixConstraintAppendixNUAppendixDIAG})
\begin{align}
  &
  \sum\limits_{j,j^{\prime}=0}^{D-1}\sum\limits_{k,k^{\prime}=0}^{n-1}
             u_{jk}S_{jk;j^{\prime}k^{\prime}}u^*_{j^{\prime}k^{\prime}}
             +
   \sum\limits_{j,j^{\prime}=0}^{D-1}        
   \lambda_{jj^{\prime}}\left[\delta_{jj^{\prime}}-\sum\limits_{k^{\prime}=0}^{n-1}u_{jk^{\prime}} u^*_{j^{\prime}k^{\prime}} \right]
   \xrightarrow[u]{\quad }\max
   \label{lagrangetovariateNUDlen}
\end{align}
Despite the matrix $u_{jk}$ being real we write it in a ``complex'' form
to variate separately over $u_{jk}$ and $u^*_{jk}$.
The tensor $S_{jk;j^{\prime}k^{\prime}}=S^*_{j^{\prime}k^{\prime};jk}$ is Hermitian.
The variations
\begin{subequations}
  \label{variatelagrangetovariate}
\begin{align}
 0&= \sum\limits_{j^{\prime}=0}^{D-1}\sum\limits_{k^{\prime}=0}^{n-1}u_{j^{\prime}k^{\prime}}S_{j^{\prime}k^{\prime};jk}
 -
\sum\limits_{j^{\prime}=0}^{D-1}
\lambda_{j^{\prime}j}u_{j^{\prime}k} \label{variationUstar}\\
 0&= \sum\limits_{j^{\prime}=0}^{D-1}\sum\limits_{k^{\prime}=0}^{n-1}
 S_{jk;j^{\prime}k^{\prime}}u^*_{j^{\prime}k^{\prime}}
 -\sum\limits_{j^{\prime}=0}^{D-1}\lambda_{jj^{\prime}}u^*_{j^{\prime}k}
\end{align}
\end{subequations}
are consistent only when $\lambda_{jj^{\prime}}$
is a Hermitian matrix
\begin{align}
  \lambda_{jj^{\prime}}&=\lambda^*_{j^{\prime}j}
  \label{lamdaHermitian}
\end{align}
From (\ref{variatelagrangetovariate}) it follows
that 
the functional (\ref{allProjUKxfAppendixDIAG}) extremal value is equal
to the spur of Lagrange multipliers matrix $\lambda_{jj^{\prime}}$:
\begin{align}
  \max{\mathcal F}&=\sum\limits_{j=0}^{D-1}\lambda_{jj}
  \label{Fextremal}
\end{align}

An iteration algorithm finding the maximum of  (\ref{allProjUKxfAppendixDIAG})
subject to (\ref{optimmatrixConstraintAppendixNUAppendixDIAG}) constraints is:
\begin{enumerate}
\item
\label{firstStepLambda}
  Take initial $\lambda_{ij}$ and solve  optimization problem (\ref{lagrangetovariateNUDlen})
  with respect to $u_{jk}$
subject to partial constraint (\ref{optimmatrixConstraintScalarAppendix}).
Solution method --- an eigenvalue problem of $Dn$ dimension
  in a vector space
  formed by writing all $u_{jk}$ matrix elements
  in a vector,
   row by row.
   The result: $p=0\dots Dn-1$ eigenvalues ${\mathcal F}^{[p]}$ and corresponding
    matrices $u_{jk}^{[p]}$ reconstructed
   back from the eigenvectors, row by row.
\item
\label{EVSelectionFromAll}
To select the $u_{jk}$  among all $Dn$ eigenstates one need to try a number of them,
selecting the ones providing a large value of the original
functional.
Taking only the state of the maximal eigenvalue
typically gives a local maximum.
Chosen $u_{jk}$
 is not partially unitary
  as the constraint (\ref{optimmatrixConstraintScalarAppendix}) is
  a subset of the full ones (\ref{optimmatrixConstraintAppendixNUAppendixDIAG}).
  Expand $u_{jk}$ in
\href{https://en.wikipedia.org/wiki/Singular_value_decomposition}{SVD}:
  \begin{align}
    u_{jk}&=\sum\limits_{j^{\prime}=0}^{D-1}\sum\limits_{k^{\prime}=0}^{n-1}
    U_{jj^{\prime}}\Sigma_{j^{\prime}k^{\prime}}V^{\dagger}_{k^{\prime}k}
    \label{SVDNUDlen}
  \end{align}
  and adjust all SVD numbers to $\pm 1$. The $\Sigma_{jk}=\delta_{jk}$
  is typically the best option as this is the minimal change
  (initial $\Sigma_{jj}$ are positive).
  Obtained
\begin{align}
\widetilde{u}_{jk}=\sum\limits_{s=0}^{ \min(D,n)-1}
    U_{js}V^{\dagger}_{sk}
    \label{SVDadjustmentSimple}
\end{align} 
  is a partially unitary matrix
  satisfying all the constraints (\ref{optimmatrixConstraintAppendixNUAppendixDIAG}).
  This $\widetilde{u}_{jk}$ becomes the next iteration $u_{jk}$
  of the solution.
  Because of ${u}_{jk}\to\widetilde{u}_{jk}$ adjustment 
  the value of ${\mathcal F}$ becomes less optimal.
  There are other methods to adjust the $u_{jk}$
  to satisfy the full set of 
  (\ref{optimmatrixConstraintAppendixNUAppendixDIAG}) constraints,
  for example an eigenvector expansion of the
  matrix $\sum_{k=0}^{n-1}u_{jk}u_{j^{\prime}k}$ followed by
  eigenvalues adjustment\cite{gsmalyshkin2017comparative},
  Gram--Schmidt orthogonalization, etc.
  However, the SVD expansion (\ref{SVDNUDlen}) is special,
  see (\ref{SVDSpecialRole}) below.  
\item \label{lagrangeMultipliersStep} Put this new $u_{jk}$ to (\ref{variationUstar}),
  then multiply it by $u^*_{ik}$ and sum over $k=0\dots n-1$.
  As the $u_{jk}$ is partially unitary  (\ref{optimmatrixConstraintAppendixNUAppendixDIAG})
  obtain new values for Lagrange multipliers $\widetilde{\lambda}_{ij}$
  and take it's Hermitian part\footnote{The equation for Lagrange multipliers (\ref{newLambdaSolPartial}) produces
  an arbitrary matrix $\widetilde{\lambda}_{ij}$;
a variation of the constraints produces Hermitian matrix
$\lambda_{ij}$.
Lagrange multipliers in (\ref{variatelagrangetovariate})
should be set to make the first variation at given  $u_{jk}$
as close to zero as possible;
least squares expansion of the first variation ($D\times n$ matrix)
in Lagrange multipliers ($D\times D$ matrix) gives (\ref{newLambdaSolPartial}).
For an arbitrary matrix $\mathcal{A}$ it's best approximation by
a Hermitian matrix $\mathcal{B}$ is the Hermitian part
$\mathcal{B}=\mathcal{A}_H=
\frac{1}{2}(\mathcal{A}+\mathcal{A}^{\dagger})$.
This follows immediately from the Frobenius norm triangle inequality
by splitting the matrix into
\href{https://en.wikipedia.org/wiki/Hermitian_matrix}{Hermitian $\mathcal{A}_H$}
and
\href{https://en.wikipedia.org/wiki/Skew-Hermitian_matrix}{anti--Hermitian $\mathcal{A}_{AH}$}
 parts:
$\|\mathcal{B}-\mathcal{A}\|_F=
\|\mathcal{B}-\mathcal{A}_H-\mathcal{A}_{AH}\|_F\le
\|\mathcal{B}-\mathcal{A}_H\|_F+
\|\mathcal{A}_{AH}\|_F$.
}:
  \begin{align}
    \widetilde{\lambda}_{ij}&=
    \sum\limits_{j^{\prime}=0}^{D-1}\sum\limits_{k,k^{\prime}=0}^{n-1}u_{j^{\prime}k^{\prime}}S_{j^{\prime}k^{\prime};jk}u^*_{ik}
    \label{newLambdaSolPartial} \\    
    \lambda_{ij}&=\frac{1}{2}
    \left[\widetilde{\lambda}_{ij}+\widetilde{\lambda}^*_{ji}\right]
    & i,j=0\dots D-1
    \label{newLambdaPartial}    
  \end{align}
  This  $\lambda_{ij}$ is 
  the next iteration of Lagrange multipliers.
  As iterations proceed -- the  $\widetilde{\lambda}_{ij}$
  is expected to converge to a Hermitian matrix by itself,
  without (\ref{newLambdaPartial}) required.
  For original (not yet full--constraint adjusted) $u_{jk}$, which is an eigenvector of $S_{j^{\prime}k^{\prime};jk}$,
  the $\widetilde{\lambda}_{ij}$ is Hermitian.
  The anti--Hermitian part of $\widetilde{\lambda}_{ij}$
  cancels in the quadratic form (\ref{lagrangetovariateNUDlen}).
  One can possibly obtain a Hermitian $\lambda_{ij}$ right away with multiplication
  of (\ref{variationUstar}) by itself (instead of $u_{jk}$ for (\ref{newLambdaSolPartial}));
  the Hermitian $\lambda_{ij}$ is then obtained from $\lambda_{ij}^2$
  as all the eigenvalues of $\lambda_{ij}$ are all positive;
  the result is very similar to (\ref{newLambdaPartial}),
  a drawback for this new $\lambda_{ij}$
  --- the (\ref{Fextremal}) now holds only approximately for current
  iteration
of $u_{jk}$,
see
  \texttt{\seqsplit{com/polytechnik/utils/KGOIterationalLambda2.java}}.
  \item
    Put this new $\lambda_{ij}$ to (\ref{lagrangetovariateNUDlen}) and repeat iteration process until converged.
    On the first iteration take initial values of Lagrange multipliers $\lambda_{ij}=0$.
\end{enumerate}
For a simpler scalar QCQP optimization problem
of 
 \cite{MalMuseScalp}, ``\emph{Appendix F: Directional Information:
  $I\xrightarrow[{\psi}]{\quad }\max$
  Subject To the  Constraint $\Braket{\psi|C|\psi}=0$}'',
where we considered a single quadratic constraint,
similar iteration algorithm
converges fast but may 
fail when optimization and constraint
matrices  have a number of eigenvectors in common.
The optimization problem
(\ref{allProjUKxfAppendixDIAG})
subject to (\ref{optimmatrixConstraintAppendixNUAppendixDIAG}) constraints
is a problem of (\ref{eigenvaluesLikeProblem}) type,
it has a more complex internal structure
than the problem considered in \cite{MalMuseScalp}.

The described Lagrange multipliers algorithm is based on eigenvalue problem
solution: (\ref{lagrangetovariateNUDlen}) with
partial constraint (\ref{optimmatrixConstraintScalarAppendix})
as normalizing: $D=\Braket{\psi^2}$.
It is much less sensitive
to degeneracy than
\href{https://en.wikipedia.org/wiki/Newton%27s_method#k_variables,_k_functions}{Newtonian type}
iterations, where even a single
degenerate degree of freedom
makes linear system 
(with \href{https://en.wikipedia.org/wiki/Hessian_matrix}{Hessian matrix})
iteration to fail.
A question arise when the described above iteration algorithm fails.
Currently --- we do not have the exact answer;
the condition of iteration algorithm
convergence requires a separate study.
The  algorithm  does not converge well for
partially unitary operators with $D<n$, but given large enough
iterations number it produces a good enough solution.
The reason for a slow convergence is that with (\ref{newLambdaSolPartial})
$\lambda_{ij}$ the Hessian matrix is degenerated at
the adjusted $u_{jk}$ (\ref{SVDadjustmentSimple}) ---
at this $u_{jk}$ not only first but also second variation of
the objective function is zero;
this is a
\href{https://en.wikipedia.org/wiki/Lagrange_multiplier#Multiple_constraints}{constraint qualification} problem.
The algorithm does not diverge, it provides a sequence of close to optimal solutions.
See \texttt{\seqsplit{com/polytechnik/utils/KGOIterationalSimpleOptimizationU.java}}
for a numerical implementation.
We also tried to find an algorithm
of 
\href{https://en.wikipedia.org/wiki/Contraction_mapping}{contraction mapping}
type, but this requires more study.
The convergence can be greatly improved using
linear constraints, see Appendix \ref{linearConstraintOpt}
below where the constraints (\ref{optimmatrixConstraintAppendixNUAppendixDIAG})
were replaces  by the closeness of $u_{jk}$
to current iteration value (\ref{constraintsCloseness}).
In many situation, however, an approximate solution
is sufficient.

\subsection{\label{SBDbasisForAdjustment}On Constrained Optimization In The Singular Values Basis}
Before we go further let us discuss the roles of
(\ref{SVDNUDlen}) singular values
and their relation to the calculation of Lagrange multipliers.
If we write optimization problem (\ref{lagrangetovariateNUDlen})
in SVD basis (\ref{SVDNUDlen})
the $u_{jk}$ is represented as
a product of three matrices.
The constraints
(\ref{optimmatrixConstraintAppendixNUAppendixDIAG})
require  all singular values $\Sigma_{jj}=\pm 1$.
We denote this diagonal matrix as vector $\Sigma_{s}$.
The objective function  (\ref{allProjUKxfAppendix})
is then $\mathcal{F}=\sum_{s,s^{\prime}=0}^{D-1}
\Sigma_{s} \widetilde{S}_{ss^{\prime}} \Sigma_{s^{\prime}}$.
Obtain constrained optimization problem with $D$ Lagrange
multipliers $\widetilde{\lambda}_s$
\begin{align}
&\sum\limits_{s,s^{\prime}=0}^{D-1}
\Sigma_{s} \widetilde{S}_{ss^{\prime}} \Sigma_{s^{\prime}} +\sum\limits_{s=0}^{D-1}
\widetilde{\lambda}_s\left[1-\Sigma_s^2\right]\to\max
\label{optimizationProblemSigmaBasis} \\
&\widetilde{S}_{ss^{\prime}}=
 \sum\limits_{j,j^{\prime}=0}^{D-1}\sum\limits_{k,k^{\prime}=0}^{n-1}
 U_{js}V^{\dagger}_{sk}
  S_{jk;j^{\prime}k^{\prime}}
U_{j^{\prime}s^{\prime}}V^{\dagger}_{s^{\prime}k^{\prime}}
\label{SinSigma}
\end{align}
from which we immediately obtain the values
\begin{align}
\widetilde{\lambda}_s&=\frac{1}{\Sigma_{s}} \sum\limits_{s^{\prime}=0}^{D-1}
\widetilde{S}_{ss^{\prime}}\Sigma_{s^{\prime}} 
\label{lambdaNotYetAdjustedSigma}
\end{align}
for all  adjusted $\Sigma_{s}=1$
\begin{align}
\widetilde{\lambda}_s&= \sum\limits_{s^{\prime}=0}^{D-1}
\widetilde{S}_{ss^{\prime}}
\label{lambdaAdjustedSigma}
\end{align}
Comparing (\ref{optimizationProblemSigmaBasis})
with (\ref{lagrangetovariateNUDlen})
obtain $\lambda_{ij}$ in original basis
\begin{align}
\widetilde{\lambda}_s&= \sum\limits_{i,j=0}^{D-1}
\lambda_{ij}U_{is}U_{js}
\label{lambdaAdjustedSigmaConversionB1} \\
\lambda_{ij}&=\sum\limits_{s=0}^{D-1}
U_{is}U_{js}\widetilde{\lambda}_s
\label{lambdaHopefullyConvert}
\end{align}
Whereas the original functional (\ref{lagrangetovariateNUDlen})
has $D^2$ Lagrange multipliers $\lambda_{ij}$,
the (\ref{optimizationProblemSigmaBasis}) has only $D$ ---
a constraint for every singular value of the matrix $u_{jk}$;
it is clear why: since
the partial constraint (\ref{optimmatrixConstraintScalarAppendix})
is always satisfies from the eigenproblem
it is sufficient to set $D-1$
diagonal elements of (\ref{optimmatrixConstraintAppendixNUAppendixDIAG}) to $1$, then all off--diagonal elements
are immediately zero.

\subsection{\label{StepWithoitSVD}On Iteration Step Without Using The SVD}
In the algorithm above we extensively used SVD expansion (\ref{SVDNUDlen})
for iterations.
Let us consider
how to avoid using the SVD by replacing it
with an eigenvalue problem of the dimension $D\times D$
for the purpose of both: computational complexity and better understanding of the algorithm.
Obtained partial constraint (\ref{optimmatrixConstraintScalarAppendix})
solution matrix $u_{jk}$
is non--orthogonal, the Gram matrix is:
\begin{align}
G^{u}_{jj^{\prime}}&=\sum\limits_{k=0}^{n-1}u_{jk}u_{j^{\prime}k}
\label{GramUpartial}
\end{align}
We need to ``adjust'' $u_{jk}$ to satisfy the full set
of  (\ref{optimmatrixConstraintAppendixNUAppendixDIAG}) constraints.
Consider the eigenstates of the Gram matrix
\begin{align}
\Ket{G^{u}|u^{[i]}}&=\lambda_G^{[i]}\Ket{u^{[i]}}
\label{GramMatrixEV}
\end{align}
The eigenvalues of this problem are equal to the singular values (\ref{SVDNUDlen})
squared $\lambda_G^{[i]}=\Sigma^2_{ii}$.
Setting all $\lambda_G^{[i]}=1$
(eigenvalues adjustment technique \cite{gsmalyshkin2017comparative})
produces a new basis in which (\ref{optimmatrixConstraintAppendixNUAppendixDIAG})
constraints are satisfied in full.
The result is identical to the transform (\ref{SVDadjustmentSimple})
of setting all $\Sigma_{jj}=1$
but it is obtained without solving a SVD problem, the
eigenvalue  $D\times D$
problem (\ref{GramMatrixEV}) is used instead,
 see
\texttt{\seqsplit{com/polytechnik/utils/KGOEVSelection.java:getEVAdjustedTo1()}} for
an implementation.

Optimization problem is question is invariant relatively a unitary transform
(the $A_{sj}$ is a unitary matrix)
\begin{align}
v_{sk}&=\sum\limits_{j=0}^{D-1}A_{sj}u_{jk} \label{Vdef}
\end{align}
The tensor $S_{jk;j^{\prime}k^{\prime}}$ transforms with $A_{sj}$ as (\ref{Stransform}),
Gram matrix (\ref{GramUpartial})
corresponds to the tensor $S_{jk;j^{\prime}k^{\prime}}=G^{u}_{jj^{\prime}}\delta_{kk^{\prime}}$.
\begin{align}
S_{sk;s^{\prime}k^{\prime}}&=
\sum\limits_{jj^{\prime}=0}^{D-1} A_{sj} S_{jk;j^{\prime}k^{\prime}} A_{s^{\prime}j^{\prime}}
\label{Stransform}\\
\mathcal{F}&=\sum\limits_{j,j^{\prime}=0}^{D-1}\sum\limits_{k,k^{\prime}=0}^{n-1}
u_{jk}S_{jk;j^{\prime}k^{\prime}}u_{j^{\prime}k^{\prime}}=
\sum\limits_{s,s^{\prime}=0}^{D-1}\sum\limits_{k,k^{\prime}=0}^{n-1}
v_{sk}S_{sk;s^{\prime}k^{\prime}}v_{s^{\prime}k^{\prime}}
\label{finv}
\end{align}
The constraints
for new variables $v_{sk}$ have the same form (\ref{optimmatrixConstraintAppendixNUAppendixDIAG})
\begin{align}
\delta_{ss^{\prime}}&=\sum\limits_{k,k^{\prime}=0}^{n-1}v_{sk}v_{s^{\prime}k} & s,s^{\prime}=0\dots D-1
\label{VBasisconstraintDirect}
\end{align}
Let us transform the input to
the basis of  Gram matrix eigenvectors.
Solve generalized eigenproblem (\ref{GramMatrixEV})
to find the eigenvalues $\lambda_G^{[s]}$
  and the eigenvectors $v^{[s]}_{j}$ of the Gram matrix $G^{u}_{jj^{\prime}}$
  \begin{align}
  \sum\limits_{j^{\prime}=0}^{D-1}G^{u}_{jj^{\prime}}
  v^{[s]}_{j^{\prime}}&=\lambda_G^{[s]} v^{[s]}_{j}
  \label{GramEVtransform}
  \end{align}
Were it all $\lambda_G^{[s]}=1$ --- the eigenstates of
the Gram matrix would form the sought partially unitary operator, but this
is typically not.
Take Gram matrix eigenvectors as a new basis, the unitary transform matrix
is $A_{sj}=v^{[s]}_{j}$, and write optimization
problem (\ref{finv}) in this
new basis $v_{sk}$ (\ref{Vdef})
with the tensor $S_{sk;s^{\prime}k^{\prime}}$
transformed from the $S_{jk;j^{\prime}k^{\prime}}$
according to (\ref{Stransform}).
 If all scaling coefficients $\mu_s=1$ --- this would be
exactly the original problem since it is invariant relatively
unitary transforms of the basis,
but if we put the
factors  $\mu_s$ (\ref{Mu_S}) --- this
makes the solution to satisfy (\ref{VBasisconstraintDirect});
non--unitary scaling factors $\mu_s$ adjust the solution to satisfy the full set of the constraints.
\begin{align}
&\mu_s=\pm\frac{1}{\sqrt{\lambda_G^{[s]}}} \label{Mu_S} \\
&\sum\limits_{s,s^{\prime}=0}^{D-1}\sum\limits_{k,k^{\prime}=0}^{n-1}
             \mu_s v_{sk}S_{sk;s^{\prime}k^{\prime}}v_{s^{\prime}k^{\prime}}\mu_{s^{\prime}}
    \xrightarrow[v]{\quad }\max
   \label{lagrangetovariateNUDMuScaleV}   
\end{align}
This scaling adjustment performed in Gram matrix basis
is an alternative to SVD adjustment (\ref{SVDadjustmentSimple}).
One need to convert the problem from original basis
to the basis of Gram matrix eigenvectors, then scale them
by the (\ref{Mu_S}) factors.
The $\mu_s v_{sk}$ satisfies partial orthogonality constraints.
We can write optimization problem in this new basis, and perform
the iteration algorithm of Appendix \ref{numAlgorithm} above,
then ``chaining'' unitary transforms as iterations proceed,
the result will be identical as the problem is invariant relatively
these transforms, but the idea of solution adjustment in the from of pure scaling
opens a number of new ways to improve the algorithm,
see \texttt{\seqsplit{com/polytechnik/utils/KGOIterationalMultipleTransforms.java}}
for a numerical implementation.

\subsection{\label{adjustmentWithOperator}
On Operator--Dependent Solution Adjustment}

In the previous section we considered solution adjustment procedure
applied to some initial ``partial'' solution.
This adjustment is a non--unitary basis transform.
A question arise about a generalization: applying
some other non--unitary transform before the adjustment.
Optimization problem in question is to maximize (\ref{lagrangetovariateNUDMuScaleVInitial})
subject to (\ref{VBasisconstraintDirectInitial}) constraints:
\begin{align}
\frac{\mathcal{F}}{D}=&\frac{
\sum\limits_{j,j^{\prime}=0}^{D-1}\sum\limits_{k,k^{\prime}=0}^{n-1}
             u_{jk}S_{jk;j^{\prime}k^{\prime}}u_{j^{\prime}k^{\prime}}
             }{
             \sum\limits_{j=0}^{D-1}\sum\limits_{k=0}^{n-1}
             u^2_{jk}
             }
    \xrightarrow[u]{\quad }\max    
   \label{lagrangetovariateNUDMuScaleVInitial}   \\
   \delta_{jj^{\prime}}&=\sum\limits_{k,k^{\prime}=0}^{n-1}u_{jk}u_{j^{\prime}k} & j,j^{\prime}=0\dots D-1
\label{VBasisconstraintDirectInitial}
\end{align}
Consider a Hermitian operator $\mathcal{J}$ with matrix
elements $\mathcal{J}_{jj^{\prime}}$, this can be e.g.
Lagrange multipliers  matrix (\ref{newLambdaPartial}), unit matrix, etc.
A generalized eigenvalue problem with
$\mathcal{J}_{jj^{\prime}}$ and $G^{u}_{jj^{\prime}}$ (\ref{GramUpartial})
matrices is formulated as
\begin{align}
  \sum\limits_{j^{\prime}=0}^{D-1}\mathcal{J}_{jj^{\prime}}
  v^{[s]}_{j^{\prime}}
  &=
  \lambda_{\mathcal{J}}^{[s]}
  \sum\limits_{j^{\prime}=0}^{D-1}G^{u}_{jj^{\prime}}
  v^{[s]}_{j^{\prime}}
    \label{JGramEVtransform}
\end{align}
Because of the Gram matrix $G^{u}_{jj^{\prime}}$ in the
right hand side obtained solution
\begin{align}
v_{sk}&=\sum\limits_{j=0}^{D-1}v^{[s]}_{j} u_{jk}
\label{SomVfromUJG}
\end{align}
satisfies (\ref{VBasisconstraintDirectInitial})
constraints 
$\delta_{ss^{\prime}}=\sum_{k,k^{\prime}=0}^{n-1}v_{sk}v_{s^{\prime}k}$.
The transform $v^{[s]}_{j}$ is non--unitary
\begin{subequations}
\label{nonUnitaryTransform}
\begin{align}
\delta_{ss^{\prime}}&=
\sum\limits_{j,j^{\prime}=0}^{D-1}
v^{[s]}_{j} G^{u}_{jj^{\prime}} v^{[s^{\prime}]}_{j^{\prime}}
\label{VGdiag} \\
G^{u;-1}_{jj^{\prime}}&=
\sum\limits_{s=0}^{D-1} v^{[s]}_{j} v^{[s]}_{j^{\prime}}
\label{VGSumS}
\end{align}
\end{subequations}
Condition (\ref{VGdiag}) creates the basis (\ref{SomVfromUJG})
satisfying partial orthogonality constraints.
Let us write the optimization problem
(\ref{lagrangetovariateNUDMuScaleVInitial}) in
this new basis $v_{sk}$. Using
\begin{align}
u_{jk}&=\sum\limits_{j^{\prime},s=0}^{D-1}
G^{u}_{jj^{\prime}} v^{[s]}_{j^{\prime}}
 v_{sk}
\label{SomUfromVJG}
\end{align}
obtain the original problem
(\ref{lagrangetovariateNUDMuScaleVInitial}) with the tensor
$S_{sk;s^{\prime}k^{\prime}}$ instead  of $S_{jk;j^{\prime}k^{\prime}} $
\begin{align}
S_{sk;s^{\prime}k^{\prime}} &=
\sum\limits_{j,j^{\prime},i,i^{\prime}=0}^{D-1}
v^{[s]}_{i} G^{u}_{ij}  S_{jk;j^{\prime}k^{\prime}}
              G^{u}_{i^{\prime}j^{\prime}}
              v^{[s^{\prime}]}_{i^{\prime}}
  \label{allTheSAmeS}
\end{align}
This is a generalization of (\ref{Stransform}) to non--unitary transforms.
This is exactly the original problem (without an adjustment), but written
in the $v_{sk}$ basis.

It can be noticed that adjustment procedure of
previous section is actually a non--unitary transform with the
inverse square root of the Gram matrix $G^{u;-1/2}_{jj^{\prime}}$
(\ref{Mu_S}); there are $2^{D-1}$ distinct combinations of signs
but we take all equal to $1$.
The adjustment is equivalent to multiplying (\ref{SomUfromVJG})
by $G^{u;-1/2}_{jj^{\prime}}$ to obtain the ``adjusted'' tensor
\begin{align}
S^{adj}_{sk;s^{\prime}k^{\prime}} &=
\sum\limits_{j,j^{\prime},i,i^{\prime}=0}^{D-1}
v^{[s]}_{i} G^{u;1/2}_{ij}  S_{jk;j^{\prime}k^{\prime}}
              G^{u;1/2}_{i^{\prime}j^{\prime}}
              v^{[s^{\prime}]}_{i^{\prime}}
  \label{allTheSAmeSAdj}
\end{align}
This way the adjustment is ``transferred'' from the state $u_{jk}$
to operator $S_{sk;s^{\prime}k^{\prime}}$. Equivalent $u_{jk}$ adjustment
corresponds to $u_{jk}=\sum_{j^{\prime},s=0}^{D-1}
G^{u;1/2}_{jj^{\prime}} v^{[s]}_{j^{\prime}}
 v_{sk}$. The (\ref{allTheSAmeSAdj}) is an important option to transfer
an adjustment from a state to tensor, this allows to combine
the adjustment with optimization algorithm.
Considered in Section \ref{StepWithoitSVD} above adjustment
procedure corresponds to
 $\mathcal{J}_{jj^{\prime}}$
 being a unit matrix.

\subsection{\label{linearConstraintOpt}
On Optimization Algorithm With Linear Constraint Iteration}
In previous sections we considered
optimization algorithm with quadratic constraints
of (\ref{optimmatrixConstraintAppendixNUAppendixDIAG}) form.
In numerical implementation (\ref{lagrangetovariateNUDlen}) these constraints
lead to a poor convergence since at the point
the constraints are applied the Hessian matrix is degenerated.
Consider a linear type of constraints.

Extend $u_{jk}$ with one more degree of freedom $\chi$
to form a ``vector'' of the  dimension $D\times n+1$.
\begin{align}
\bm{z}&=\begin{pmatrix}
u_{jk}\\
\chi
\end{pmatrix}
\label{zdef}
\end{align}
Then the quadratic from
\begin{align}
\mathcal{F}&=
\frac{\bm{z}^T \mathcal{S} \bm{z}}{\bm{z}^T  \bm{z}}=
\frac{
\chi^2S_0
+2\chi\sum\limits_{j=0}^{D-1}\sum\limits_{k=0}^{n-1} b_{jk}u_{jk}
+\sum\limits_{j,j^{\prime}=0}^{D-1}\sum\limits_{k,k^{\prime}=0}^{n-1}
             u_{jk}S_{jk;j^{\prime}k^{\prime}}u_{j^{\prime}k^{\prime}} }
             {
             \chi^2+\sum\limits_{j=0}^{D-1}\sum\limits_{k=0}^{n-1} u^2_{jk}
             }
\label{finZ}
\end{align}
has the matrix $\mathcal{S}$
\begin{align}
\mathcal{S}&=
\left(
\begin{array}{c|c}
S_{jk;j^{\prime}k^{\prime}} & b_{j^{\prime}k^{\prime}} \\
\hline
b_{jk} & S_0
\end{array}
\right)
\label{SmatrixNZ}
\end{align}
The idea is to consider the $b_{jk}$ and $S_0$ as some
kind of ``Lagrange Multipliers'' to set the variation of (\ref{finZ})
to zero at the ``adjusted'' $u_{jk}$,
denote it as iteration value $u^{IT}_{jk}$.
Consider the constraints
\begin{align}
u_{jk}-u^{IT}_{jk}=0
\label{constraintsCloseness}
\end{align}
this is the closeness of $u_{jk}$ to current iteration value $u^{IT}_{jk}$
(adjusted value satisfying all the
required constraints (\ref{optimmatrixConstraintAppendixNUAppendixDIAG})).
A one more degree of freedom $\chi$ was introduced
to preserve the form of the
\href{https://en.wikipedia.org/wiki/Rayleigh_quotient}{Rayleigh quotient}
for the optimization problem (\ref{finZ}).
Variating it over $u_{jk}$ and $\chi$
obtain (\ref{varujk}) and (\ref{varchi}) respectively;
in these formulas $u_{jk}=u^{IT}_{jk}$,
$F_0^{IT}$ is a known constant, and $B_0$ and $S_0$ are unknown constants.
\begin{align}
F_0^{IT}&=\sum\limits_{j,j^{\prime}=0}^{D-1}\sum\limits_{k,k^{\prime}=0}^{n-1}
             u_{jk}S_{jk;j^{\prime}k^{\prime}}u_{j^{\prime}k^{\prime}}
             \label{F0IT} \\
B_0&=\sum\limits_{j=0}^{D-1}\sum\limits_{k=0}^{n-1}
b_{jk}u_{jk} \label{B0IT} \\
0&=\sum\limits_{j^{\prime}=0}^{D-1}\sum\limits_{k^{\prime}=0}^{n-1}
             S_{jk;j^{\prime}k^{\prime}}u_{j^{\prime}k^{\prime}}
+\chi b_{jk} - \frac{\chi^2 S_0 +2\chi B_0 +F_0^{IT}}{D+\chi^2} u_{jk}
             \label{varujk}\\
0&=\chi S_0 + B_0 - \frac{\chi^2 S_0 +2\chi B_0 +F_0^{IT} }{D+\chi^2} \chi
\label{varchi} \\
0&=F_0^{IT}+\chi B_0
-\left(\chi^2 S_0 +2\chi B_0 +F_0^{IT}\right)\frac{D}{D+\chi^2}
\label{varujkSumjk}
\end{align}
Multiply (\ref{varujk}) by $u_{jk}$ and sum it over $j$ and $k$, obtain
(\ref{varujkSumjk}). For a given $\chi$ the (\ref{varchi}) and (\ref{varujkSumjk})
can be considered as a linear system for $B_0$ and $S_0$.
Obtained $2\times 2$ linear system is degenerated and has multiple solutions:
\begin{align}
(D-\chi^2) B_0
+
(D\chi)  S_0 &=
F_0^{IT} \chi
\label{B0S0_EQ1} 
\end{align}
The specific set $(\chi,S_0,B_0)$  should be selected for best convergence.
The selection
\begin{subequations}
\label{simpleSel}
\begin{align}
\chi&=1 \label{chiConstraint1}\\
S_0&=F_0^{IT}\\
B_0&=-S_0
\end{align}
\end{subequations}
is the first one to try.

\begin{enumerate}
\item
Take the values of $b_{jk}$ and $S_0$
to construct (\ref{SmatrixNZ}).
\item
Solve (\ref{finZ}) and
\hyperref[EVSelectionFromAll]{select} the most appropriate vector $\mathbf{z}$.
The result of this step --- the ``adjusted'' $u^{IT}_{jk}$ satisfying
all the required constraints (\ref{optimmatrixConstraintAppendixNUAppendixDIAG}).
\item
Take this new $u^{IT}_{jk}$,
and select some value of  $\chi$, for example (\ref{simpleSel}),
calculate  ``Lagrange Multipliers'' $b_{jk}$ (\ref{varujk}) and $S_0$ (\ref{B0S0_EQ1})
to construct (\ref{SmatrixNZ}) matrix.
If one uses $\chi$ value from (\ref{finZ}) maximization problem ---
iterations typically stick to some local maximum.
If one uses a fixed value for $\chi$, such as (\ref{chiConstraint1})
--- a convergence
is observed;
not very fast, but better than in the Appendix \ref{numAlgorithm} above.
Repeat iteration process. On the first iteration take $b_{jk}=S_0=0$.
\end{enumerate}
This ``Linear constraints'' algorithm is implemented in the
\texttt{\seqsplit{com/polytechnik/utils/KGOIterationalLinearConstraintsE.java}}.
An attempt to use $D$ extra degrees of freedom instead of a single
one was much less successful
\texttt{\seqsplit{com/polytechnik/utils/KGOIterationalLinearConstraintsExtraDegreesOfFreedom.java}}.

\subsection{\label{numAlgorithmApproximate}
An Algorithm to Find an Approximate
Solution to
the Knowledge Generalizing Operator}

Consider the same problem
(\ref{allProjUKxfAppendixDIAG})
subject to (\ref{optimmatrixConstraintAppendixNUAppendixDIAG})
constraint.
The bases are considered already orthogonalized:
$\delta_{jj^{\prime}}=\Braket{f_j|f_{j^{\prime}}}$
and $\delta_{kk^{\prime}}=\Braket{x_k|x_{k^{\prime}}}$.
Assume we found optimization problem
(\ref{allProjUKxfAppendixDIAG})
solution
with ``partial'' constraints (\ref{optimmatrixConstraintScalarAppendix}),
this is (\ref{lagrangetovariateNUDlen}) with $\lambda_{ij}=0$.
Put it to (\ref{optimmatrixConstraintAppendixNUAppendix}) and expand $u_{jk}$ in SVD:
\begin{align}
 u_{jk}&=\sum\limits_{j^{\prime}=0}^{D-1}\sum\limits_{k^{\prime}=0}^{n-1}
    U_{jj^{\prime}}\Sigma_{j^{\prime}k^{\prime}}V^{\dagger}_{k^{\prime}k}
    \label{SVDSpecialRoleExpansion} \\
 \Braket{f_j|f_{j^{\prime}}}&= \sum\limits_{k,k^{\prime} =0}^{n-1}u_{jk}
\Braket{x_k|x_{k^{\prime}}}
u^*_{j^{\prime}k^{\prime}}& j,j^{\prime}=0\dots D-1
    \label{optimmatrixConstraintAppendixNUASSVD}
\end{align}
Write (\ref{optimmatrixConstraintAppendixNUASSVD})
for orthogonal bases $\mathfrak{x}_k$ and $\mathfrak{f}_j$
\begin{align}
\mathfrak{x}_k &=\sum\limits_{k^{\prime}=0}^{n-1}V^{\dagger}_{kk^{\prime}}x_{k^{\prime}} \\
\mathfrak{f}_j&=\sum\limits_{j^{\prime}=0}^{D-1}U^{\dagger}_{jj^{\prime}}f_{j^{\prime}}\\
\Braket{\mathfrak{f}_j|\mathfrak{f}_{j^{\prime}}}&=
\sum\limits_{k,k^{\prime} =0}^{n-1}
\Sigma_{jk}
\Braket{\mathfrak{x}_k|\mathfrak{x}_{k^{\prime}}}
\Sigma_{j^{\prime}k^{\prime}}& j,j^{\prime}=0\dots D-1
\label{SVDSpecialRole}
\end{align}
The (\ref{SVDSpecialRole}) is (\ref{optimmatrixConstraintAppendixNUASSVD})
written in  $\mathfrak{x}_k$ and $\mathfrak{f}_j$ orthogonal bases.
Since $\delta_{jj^{\prime}}=\Braket{\mathfrak{f}_j|\mathfrak{f}_{j^{\prime}}}$
and $\delta_{kk^{\prime}}=\Braket{\mathfrak{x}_k|\mathfrak{x}_{k^{\prime}}}$
the (\ref{SVDSpecialRole}) is satisfied
only when all singular values
of $u_{jk}$
are $\pm 1$.
Actually we made a single iteration of the algorithm above,
this $\Sigma_{jk}$-adjusted solution is an approximate solution
 one should try first.
 Since $\Sigma_{jk}$ is diagonal,
 in the $(\mathfrak{f}_j,\mathfrak{x}_{k})$ basis we have
 a one--to--one relation 
\begin{align}
\mathfrak{f}_j&=\mathfrak{x}_j\Sigma_{jj}
\label{xFnewBassis}
\end{align}
This is \textsl{not} a least squares type of relation,
for example the result is invariant
relatively the transform
$\Sigma_{jk}\to -\Sigma_{jk}$. If $(\mathfrak{f}_j,\mathfrak{x}_{k})$ basis
satisfies (\ref{SVDSpecialRole}) then all singular values are $\pm 1$
(the condition of partial unitarity)
and $(\mathfrak{f}_j,\mathfrak{x}_{k})$
relation is plain $\mathfrak{f}_j=\pm\,\mathfrak{x}_j$.
The probability (\ref{probgUypkAppAS}) in this basis is
\begin{align}
  \Braket{\psi_{\mathfrak{f}}|\mathcal{U}|\psi_{\mathfrak{x}}}^2
&=
  \frac{
 \left|
    \sum\limits_{j=0}^{D-1}
    \mathfrak{f}_j\mathfrak{x}_j\Sigma_{jj}
    \right|^2
  }
       {
         \sum\limits_{j=0}^{D-1} \mathfrak{f}^2_j
         \sum\limits_{k=0}^{n-1} \mathfrak{x}^2_k
       } \label{probgUypkAppASfbxb}
\end{align}
Partial unitarity ``adjusted'' case corresponds to $\Sigma_{jj}=\pm 1$.

Consider the meaning of a state with an arbitrary $\Sigma_{jj^{\prime}}$.
The (\ref{SVDSpecialRole}) is actually the constraint
(\ref{optimmatrixConstraintAppendixNUAppendixDIAG})
but with the positive diagonal matrix $\Sigma^2_{jj^{\prime}}$,
not $\delta_{jj^{\prime}}$.
What does it mean if we put this $u_{jk}$ ``partial constraint (\ref{optimmatrixConstraintScalarAppendix})''
solution to probability (\ref{probgUypkAppAS})
without any adjustment?
This breaks the preservation of probability,
the probability (\ref{probgUypkAppASfbxb}) is no longer $[0:1]$ bounded,
it is now
$0\le P(\mathbf{f})\Big|_{\mathbf{x}}\le \max\limits_j\Sigma^2_{jj}$;
the range  $[0:1]$ holds only ``on average'', for the entire sample.
However, this does not change the calculation of outcome value (\ref{ValueF}).
One can also modify (\ref{probgUypkAppASfbxb}) to have the probability $[0:1]$ bounded, the maximal value is $1$, it corresponds to
$\mathfrak{f}_j=\mathfrak{x}_j\Sigma_{jj}$.
\begin{align}
P(\mathfrak{f})\Big|_{\mathfrak{x}}&\approx
  \frac{
 \left|
    \sum\limits_{j=0}^{D-1}
    \mathfrak{f}_j\mathfrak{x}_j\Sigma_{jj}
    \right|^2
  }
       {
         \sum\limits_{j=0}^{D-1} \mathfrak{f}^2_j
         \sum\limits_{j=0}^{D-1} \mathfrak{x}^2_j\Sigma^2_{jj}
       } \label{probgUypkAppASfbxbApprox}
\end{align}
But this is only for evaluation,
this is not the function used in optimization problem,
optimization problem with the probability (\ref{probgUypkAppASimportantOnly})
is much more difficult.
There is a trivial option to use
the probability (\ref{probgUypkAppASimportantOnlyDegreeOfFreedom})
for optimization and (\ref{probgUypkAppASfbxbApprox}) for evaluation.
The $\Sigma^2_{jj}$, $j=0\dots D-1$, factor (whether the singular values are adjusted or not)
in the denominator prevents
a decrease of probability when polluting the $\mathbf{x}$--space with a large number of
completely random components (\ref{ChristoffelOriginalPhi});
the value of $\mathfrak{f}$ (\ref{ValueF})
does not depend  on this $\mathfrak{x}$--depended factor,
maximal value of probability corresponds to
$\mathfrak{f}_j=\mathfrak{x}_j\Sigma_{jj}$;
the probability is invariant with respect to $\mathfrak{f}_j\to C\mathfrak{f}_j$,
\hyperref[fcalculationFromConst]{normalize it to const}
to obtain actual values.
This partial constraint solution
of (\ref{allProjUKxfAppendixDIAG}) subject to
 (\ref{optimmatrixConstraintScalarAppendix})
is an approximate solution one may try.
Whereas a quantum channel that preserves
probability ``on average'' does not have a physical meaning,
in data analysis it is an approximation with a clear meaning:
it emphasizes (\ref{SVDSpecialRole})
internal relations with high probability,
the $\Sigma_{jj}$ factor in (\ref{probgUypkAppASfbxbApprox}).
Mathematically this means that in (\ref{eigenvaluesLikeProblem})
we allow operators
$\mathcal{U}$ that preserve Gram matrix spur:
$D=\sum\limits_{j=0}^{D-1} \Sigma^2_{jj}$,
not the Gram matrix itself (\ref{optimmatrixConstraintAppendixNUASSVD}) as previously considered;
the solution can be found from eigenproblem
(\ref{lagrangetovariateNUDlen})
in original basis with $\lambda_{ij}=0$.

Conceptually, this technique consists in taking
any approximate $u_{jk}$, such as least squares (\ref{fxapproxLS})
or any other matrix,
not necessary (\ref{lagrangetovariateNUDlen}) solution,
Gram matrix spur preservation is not required,
expanding $u_{jk}$ in SVD (\ref{SVDSpecialRoleExpansion}),
then set $\Sigma_{jj}$ to $1$ or $-1$.
There are $2^{m-1}$ distinct combinations,
typically the minimal change adjustment --- all $\Sigma_{jj}=1$ gives the best result
as the initial $\Sigma_{jj}$ are positive.
Obtained new $u_{jk}$ matrix with singular values equal to $\pm 1$ satisfies all the required constraints (\ref{optimmatrixConstraintAppendixNUASSVD}).
Alternatively one can solve the eigenproblem (\ref{GramMatrixEV})
and adjust all the eigenvectors by the factors $\pm1/\sqrt{\lambda^{[j]}}$
(\ref{Mu_S})
to obtain the same solution without using the SVD
(it is equivalent to multiplication (\ref{allTheSAmeSAdj})
of unadjusted $u_{jk}$
by inverse square root of corresponding Gram matrix).

\section{\label{PAdjusted}On Adjusted Normalizing  Of Probability}
The probability (\ref{probgUypkAppAS})
has a normalizing factor as a product of
two Christoffel functions:
on $\mathbf{x}$ and on $\mathbf{f}$
(\ref{ChristoffelfunctionsProductMoments});
these two Christoffel functions
have $n$ and $m$ degrees of freedom respectively.
In some situations it is convenient
to construct a normalizing factor where
both $\mathbf{x}$- and $\mathbf{f}$- factors
have the same number of degrees of freedom: $m$.

One can consider the probability 
adjusted to only ``important'' $\mathbf{x}$--components,
this is $\Braket{\psi_{\mathbf{g}}|\psi_{u(\mathbf{y})}}^2$
from (\ref{properlyNormalizedPsi}) expanded:
\begin{align}
  \Braket{\psi_{\mathbf{g}}|\mathcal{U}|\psi_{\mathbf{y}}}^2
&=
  \frac{
 \left|
    \sum\limits_{k=0}^{n-1}\sum\limits_{j,s=0}^{m-1}
g_{j}  G^{\mathbf{f};\,-1}_{js}
u_{sk}
y_{k}
    \right|^2
  }
       {
         \sum\limits_{j,j^{\prime}=0}^{m-1} g_{j}G^{\mathbf{f};\,-1}_{jj^{\prime}}g_{j^{\prime}}
         \sum\limits_{j,j^{\prime}=0}^{m-1}\sum\limits_{k,k^{\prime}=0}^{n-1}
         y_{k} u_{jk} G^{\mathbf{f};\,-1}_{jj^{\prime}}u_{j^{\prime}k^{\prime}} y_{k^{\prime}}
       } \label{probgUypkAppASimportantOnly}
\end{align}
Whereas this formula
for $\Braket{\psi_{\mathbf{g}}|\psi_{u(\mathbf{y})}}^2$
has a more suitable normalizing 
than (\ref{probgUypkAppAS}),
it has $u_{jk}$ in the denominator and the problem can no longer\footnote{
This difficulty does not arise with $\mathbf{x}$- and $\mathbf{f}$-
being the same space. For example for a unitary $\mathcal{U}$
the denominator does not depend on $\mathcal{U}$.
} be
reduced to the one of form (\ref{allProjUKxfAS})
that requires only the moments of Christoffel
functions product (\ref{ChristoffelfunctionsProductMoments}).
For probability evaluation, not for optimization,
this can be done straightforward (\ref{probgUypkAppASfbxbApprox}).
A quantum channel $u_{jk}$ optimizing (\ref{allProjU})
with the probability (\ref{probgUypkAppASimportantOnly})
is an interesting direction of future research,
this new problem is no longer a QCQP problem ---
it is a problem to maximize the sum of $M$ ratios
of two quadratic forms on $u_{jk}$
subject to (\ref{optimmatrixConstraintAppendixNUAS})
constraint or, more generally,
an unconstrained optimization of (\ref{properlyNormalizedPsi}).
The one in the nominator is a dyadic product squared,
the one in the denominator is non--negative,
it cancels with the nominator when it's value is close to zero.

To adjust  the number of degrees of freedom
one can use a much simpler alternative approach.
All we need is to calculate a Christoffel function in $\mathbf{x}$--space
to normalize the probability.
A trivial approach is to use the contributing subspace $\Ket{\phi^{[i]}}$,
e.g. from (\ref{phiiD}). Despite the moments $\Braket{f_jx_k}$
have been used to build the contributing subspace $\Ket{\phi^{[i]}}$,
this does not create any difficulty
as we use these projections only to construct a Christoffel function
with matched number of degrees of freedom.
The (\ref{ChristoffelAdjusted}) is invariant
with respect to $\Braket{f_jx_k}\to -\Braket{f_jx_k}$
and tends to a constant when any $\Braket{f_jx_k}\to\infty$
(factors in the denominator and  inverse $G^{fxa}_{jj^{\prime}}$ matrix).
\begin{align}
  \Braket{\psi_{\mathbf{g}}|\mathcal{U}|\psi_{\mathbf{y}}}^2
&=
  \frac{
 \left|
    \sum\limits_{k=0}^{n-1}\sum\limits_{j,s=0}^{m-1}
g_{j}  G^{\mathbf{f};\,-1}_{js}
u_{sk}
y_{k}
    \right|^2
  }
       {
         \sum\limits_{j,j^{\prime}=0}^{m-1} g_{j}G^{\mathbf{f};\,-1}_{jj^{\prime}}g_{j^{\prime}}
         \sum\limits_{i=0}^{m-1}
         {\phi^{[i]}}^2(\mathbf{y})
       } \label{probgUypkAppASimportantOnlyDegreeOfFreedom} \\
    G^{fxa}_{jj^{\prime}}&=\sum\limits_{k,k^{\prime}=0}^{n-1}
    \Braket{f_jx_k}G^{\mathbf{x};\,-1}_{kk^{\prime}} \Braket{x_{k^{\prime}}f_{j^{\prime}}}
    \label{Gfxa}\\
G^{C}_{qq^{\prime}}&=
 \sum\limits_{s,s^{\prime}=0}^{n-1}\sum\limits_{j,j^{\prime}=0}^{m-1}
    G^{\mathbf{x};\,-1}_{qs} \Braket{x_sf_j}
    G^{fxa;\,-1}_{jj^{\prime}} \Braket{f_{j^{\prime}}x_{s^{\prime}}}
    G^{\mathbf{x};\,-1}_{s^{\prime}q^{\prime}}
    \label{GcDefined}
\\
    K^{adj}(\mathbf{x})&=\frac{1}
    {
    \sum\limits_{i=0}^{m-1}
         {\phi^{[i]}}^2(\mathbf{x})
         }
         =\frac{1}{
   \sum\limits_{q,q^{\prime}=0}^{n-1}
    x_qG^{C}_{qq^{\prime}}x_{q^{\prime}}
    }
\label{ChristoffelAdjusted}
\end{align}
The value of $K^{adj}(\mathbf{x})$
is never zero on training sample since contributing subspace
always has a constant among the components.
The probability (\ref{probgUypkAppASimportantOnlyDegreeOfFreedom})
uses Christoffel function
with adjusted number of degrees of freedom
$K^{adj}(\mathbf{x})$
(\ref{ChristoffelAdjusted})
instead of the original $K(\mathbf{x})$
 (\ref{ChristoffelFunctionDef})
for the probability
(\ref{probgUypkAppAS}).
The difference between two these Christoffel functions
is in extra terms in the denominator sum.
Since the entire $\mathbf{x}$--space can be represented as
the direct sum of $\Ket{\phi^{[i]}}$ and $\Ket{\phi^{O;[i]}}$,
a subspace of $\mathbf{x}$ orthogonal to $\Ket{\phi^{[i]}}$,
the $K(\mathbf{x})$ (\ref{ChristoffelFunctionDef}) is:
\begin{align}
    K(\mathbf{x})&=\frac{1}
    {
    \sum\limits_{i=0}^{m-1}
         {\phi^{[i]}}^2(\mathbf{x})
    +\sum\limits_{i=m}^{n-1}
         {\phi^{O;[i]}}^2(\mathbf{x})        
         }
 =
    \frac{1}{\sum\limits_{k,k^{\prime}=0}^{n-1}x_kG^{\mathbf{x};\,-1}_{kk^{\prime}}x_{k^{\prime}}}
\label{ChristoffelOriginalPhi}
\end{align}
Thus we always have $K^{adj}(\mathbf{x})\ge K(\mathbf{x})$.
The moments of two Christoffel functions product are
\begin{align}
\Braket{x_{k}f_{j}|K^{adj(\mathbf{x})}K^{(\mathbf{f})}|x_{k^{\prime}}f_{j^{\prime}}}&=
\sum\limits_{l=0}^{M}\omega^{(l)}
\frac{x^{(l)}_{k}x^{(l)}_{k^{\prime}}}
{
   \sum\limits_{q,q^{\prime}=0}^{n-1}
    x_q^{(l)}G^{C}_{qq^{\prime}}x_{q^{\prime}}^{(l)}
}
\cdot
\frac{f^{(l)}_{j}f^{(l)}_{j^{\prime}}}
{
\sum\limits_{s,s^{\prime}=0}^{m-1} f^{(l)}_{s}G^{\mathbf{f};\,-1}_{ss^{\prime}}f^{(l)}_{s^{\prime}}
}
\label{ChristoffelfunctionsProductMomentsAdjusted}
\end{align}
This tensor has the same dimensions as
(\ref{ChristoffelfunctionsProductMoments}),
the difference only in normalizing ---
it uses $G^{C}_{qq^{\prime}}$ from (\ref{GcDefined}) instead of $G^{\mathbf{x};\,-1}_{qq^{\prime}}$
in (\ref{ChristoffelfunctionsProductMoments}).
Despite it now depends on $\Braket{f_jx_k}$
moments --- they are used only to construct Christoffel function
for normalizing,
this does not change the essence of the solution
due to the invariance properties of the Christoffel function.

\section{\label{SubSpaceSelection}On Contributing Subspace Selection}
Considered above optimization problem
finds partially unitary operator $u_{jk}$ that does both:
selects the contributing subspace ($m$ vectors of the dimension $n$)
and optimizes the objective function. Besides computational difficulties
this also creates a problem with normalizing since
properly normalized objective function
(\ref{properlyNormalizedPsi})
has operator $u_{jk}$ both in the nominator
and in the denominator (\ref{probgUypkAppASimportantOnly}),
thus some surrogate normalizing (\ref{SmatrixAS}),
(\ref{ChristoffelfunctionsProductMomentsAdjusted}),
(\ref{SmatrixASValue}), or (\ref{SmatrixASValueKf})
was used instead.
It is a very attractive option to split the problem into two:
\begin{itemize}
\item Find the contributing subspace $\phi^{[j]}$ of the dimension $m$.
\item Find a unitary (not partially unitary!) operator $\mathcal{U}$
mapping from $\phi^{[j]}$ space to $f_j$ space.
\end{itemize}
A simple ``projective'' example with contributing
subspace was considered in Eq. (\ref{GEVKftoXfxf}) above.
The matrix 
$K^{(\mathbf{f\to x})}_{kk^{\prime}}$ from (\ref{Kftoxsum})
has the rank at most $m$ and the best what can be obtained
in the projective paradigm 
is a solution\cite{malyshkin2019radonnikodym} of ``direct projection'' type  
where the least squares expansion $\mathbf{f}_{LS}(\mathbf{x})$ of
$\Ket{f_j}$ in $\Ket{x_k}$ (\ref{fxapproxLS})
is used 
as the localization point in (\ref{psiGlocalized})
to obtain the state
$\Ket{\psi_{\mathbf{f}_{LS}(\mathbf{x})}}$
to be used in calculation of probabilities.

Properly normalized objective function (\ref{properlyNormalizedPsi})
maximizes the probability transferred
from $\mathbf{x}$ to $\mathbf{f}$. Consider a much simpler problem:
find a subspace of $\mathbf{x}$ contributing to the coverage of $\mathbf{f}$.
The $\mathbf{f}$--coverage is determined by
$\mathbf{f}$--Christoffel function $K^{(\mathbf{f})}(\mathbf{g})$ from (\ref{ChristoffelF}).
Consider it's values in a $\psi(\mathbf{x})$ state
\begin{align}
\mathrm{Coverage}_{\psi}&=\frac{\Braket{\psi|K^{(\mathbf{f})}|\psi}}{\Braket{\psi|\psi}}
\label{psiXKf}
\end{align}
Previously we considered a similar problem where
the Christoffel function $K$ and $\psi$ both were
functions on $\mathbf{x}$,
see \cite{ArxivMalyshkinLebesgue},
\emph{Appendix B: On The Christoffel Function Spectrum}.
Now the Christoffel function is a function on $\mathbf{f}$,
and $\psi$ is a function on $\mathbf{x}$.
The (\ref{psiXKf}) can be similarly expanded in spectrum of
$\mathbf{f}$--Christoffel function matrix
\begin{align}
&\Braket{x_{k}|K^{(\mathbf{f})}|x_{k^{\prime}}}=
\sum\limits_{l=1}^{M}
\frac{x^{(l)}_{k}x^{(l)}_{k^{\prime}}}
{
\sum\limits_{j,j^{\prime}=0}^{m-1} f^{(l)}_{j}G^{\mathbf{f};\,-1}_{jj^{\prime}}f^{(l)}_{j^{\prime}}
}\omega^{(l)} \label{ChristoffelFunMomentsXX}
\end{align}
It is different from (\ref{ChristoffelFunMoments})
with $x$--moments instead of $f$--moments.
Consider generalized eigenvalue problem
\begin{align}
 \phi^{[i]}&=\sum_{k=0}^{n-1}\alpha^{\phi;[i]}_kx_k & i=0\dots n-1 ;  \label{phiiDXX}\\
 \delta_{ii^{\prime}}&=\Braket{\phi^{[i]}|\phi^{[i^{\prime}]}}=
 \sum_{k,k^{\prime}=0}^{n-1}\alpha^{\phi;[i]}_k \Braket{x_kx_{k^{\prime}}} \alpha^{\phi;[i^{\prime}]}_{k^{\prime}} \label{pisiDnormXX}\\
  \lambda^{[i]}\delta_{ii^{\prime}}&=\Braket{\phi^{[i]}|K^{(\mathbf{f})}|\phi^{[i^{\prime}]}}=
 \sum_{k,k^{\prime}=0}^{n-1}\alpha^{\phi;[i]}_k
 \Braket{x_{k}|K^{(\mathbf{f})}|x_{k^{\prime}}} \alpha^{\phi;[i^{\prime}]}_{k^{\prime}} \label{pisiDnormlamXX}\\
  \sum\limits_{k^{\prime}=0}^{n-1}& \Braket{x_{k}|K^{(\mathbf{f})}|x_{k^{\prime}}}
  \alpha^{\phi;[i]}_{k^{\prime}} =
  \lambda^{[i]} \sum\limits_{k^{\prime}=0}^{n-1} \Braket{x_kx_{k^{\prime}}}
  \alpha^{\phi;[i]}_{k^{\prime}}
\label{GEVKftoXfxfXX}
\end{align}
Because $\mathbf{x}$-- and $\mathbf{f}$-- bases are different
the condition\cite{ArxivMalyshkinLebesgue} $\Braket{1}=\sum_{i=0}^{n-1}\lambda^{[i]}$
no longer holds, it is typically $\Braket{1}\le\sum_{i=0}^{n-1}\lambda^{[i]}$
since $m\le n$;
moreover the sum of $m$ maximal eigenvalues can possibly
exceed the total weight $\Braket{1}\lesseqgtr\sum_{i=0}^{m-1}\lambda^{[i]}$.
From Christoffel function invariance it immediately
follows that
the sum of $m$ maximal eigenvalues is equal to $\Braket{1}$
if $\mathbf{f}$ and $\mathbf{x}$ belong to the same space.

The $m$ eigenstates of (\ref{GEVKftoXfxfXX})
corresponding to $m$ maximal eigenvalues $\lambda^{[i]}$, $i=0\dots n-1$
form the $m$ states contributing most to the coverage.
This is an alternative option for the contributing subspace.
The problem is now reduced to finding a unitary (not partially unitary)
operator $\mathcal{U}$ of the dimension $m\times m$ mapping
from $\phi_k$ to $f_j$, where
$\Ket{\phi}=\sum_{k=0}^{m-1}\phi_k \Ket{\phi^{[k]}}$,
\begin{align}
f_j&=\sum\limits_{k=0}^{m-1}u_{jk}\phi_k
\label{phiTof}
\end{align}
In this form the optimization problem is greatly simplified and
the $\mathbf{x}$--normalizing in (\ref{probgUypkAppASimportantOnly})
becomes $u_{jk}$ independent:
\begin{align}
  \Braket{\psi_{\mathbf{g}}|\mathcal{U}|\psi_{\mathbf{\phi}}}^2
&=
  \frac{
 \left|
    \sum\limits_{k=0}^{m-1}\sum\limits_{j,s=0}^{m-1}
g_{j}  G^{\mathbf{f};\,-1}_{js}
u_{sk}
\phi_{k}
    \right|^2
  }
       {
         \sum\limits_{j,j^{\prime}=0}^{m-1} g_{j}G^{\mathbf{f};\,-1}_{jj^{\prime}}g_{j^{\prime}}
         \sum\limits_{j,j^{\prime}=0}^{m-1} \phi_{j}G^{\phi;\,-1}_{jj^{\prime}}\phi_{j^{\prime}}
       } \label{probgUypkAppASimportantOnlyXX}
\end{align}
This probability is exactly the same as the one we considered above,
but 
with the $\phi_k$ used as the input instead of the $x_k$;
we also have $n=m$ thus the operator $u_{jk}$ is unitary!

\section{\label{Software}Software description}

\begin{itemize}
\item Install \href{https://www.oracle.com/java/technologies/javase/jdk19-archive-downloads.html}{java 19} or later.
\item Download the latest version of the source code
\href{http://www.ioffe.ru/LNEPS/malyshkin/code_polynomials_quadratures.zip}{\texttt{\seqsplit{code\_polynomials\_quadratures.zip}}}
from \cite{polynomialcode} or from
\href{https://disk.yandex.ru/d/AtPJ4a8copmZJ?locale=en}{alternative location}.

\item Decompress and recompile the program. Run a selftest.
\begin{verbatim}
unzip code_polynomials_quadratures.zip
javac -g com/polytechnik/*/*java
java com/polytechnik/utils/TestKGO
\end{verbatim}

\item Run the program with bundled deterministic
 data file, test trivial mapping.
\begin{verbatim}
java com/polytechnik/utils/KGO --data_cols=9:0,6:0,4:8:1 \
  --SKtype=FXFX_F_CHRISTOFFEL \
  --approximation=MAXEV_EVADJ \
  --data_file_to_build_model_from=dataexamples/runge_function.csv \
  --output_files_prefix=/tmp/out_
\end{verbatim}
\item
There are a number of \verb+--approximation=+ available options.
There are no perfect implementation yet available.
\end{itemize}

\bibliography{LD,mla}

\end{document}